%% file: neurips_2026.tex
\documentclass{article}

\usepackage[preprint]{neurips_2026}
\usepackage{amsmath}
\usepackage{graphicx}
\usepackage{booktabs}

\usepackage[utf8]{inputenc} 
\usepackage[T1]{fontenc}    
\usepackage{hyperref}       
\usepackage{url}            
\usepackage{booktabs}       
\usepackage{amsfonts}       
\usepackage{nicefrac}       
\usepackage[expansion=false]{microtype}      
\usepackage{xcolor}         
\usepackage{fancyvrb}       
\definecolor{readhl}{HTML}{0B6CB0}  
\newcommand{\rdhl}[1]{\textcolor{readhl}{#1}}
\usepackage{float}
\usepackage{wrapfig}        
\usepackage{multirow}
\usepackage{titletoc}       

\usepackage{custom}
\input{macros.tex}

\makeatletter
\renewcommand{\@noticestring}{$^*$Preprint. Work partially done at UW-Madison.}
\makeatother
\title{Bridging the Gap Between Latent and Explicit Reasoning with Looped Transformers}

\author{
  Ying Fan$^{1*}$ \quad Anej Svete$^{2}$ \quad Kangwook Lee$^{3,4}$ \\[6pt]
  $^{1}$Microsoft Research \quad
  $^{2}$ETH Zürich \quad
  $^{3}$KRAFTON \quad
  $^{4}$Ludo Robotics
}

\begin{document}

\maketitle

\begin{abstract}
  \input{abstract}
\end{abstract}

\input{intro}
\input{prelim}

\input{method}
\input{results}
\input{ablation}
\input{latent_analysis}
\input{conclusion}

\section*{Acknowledgments}
Anej Svete is supported by the ETH Z{\"u}rich AI Center doctoral fellowship.

\bibliography{references}
\bibliographystyle{plainnat}

\newpage
\input{appendix}

\newpage

\end{document}

%% file: macros.tex
\newcommand{\mymacro}[1]{{#1}}

\newcommand{\defn}[1]{\textbf{#1}}

\newcommand{\modelName}{\mymacro{LOTUS}\xspace}

\newcommand{\R}{{\mymacro{\mathbb{R}}}}

\DeclareMathSymbol{\mlq}{\mathord}{operators}{``} 
\DeclareMathSymbol{\mrq}{\mathord}{operators}{`'} 

\newcommand{\bigO}{{\mymacro{\mathcal{O}}}}
\newcommand{\bigOFun}[1]{{\mymacro{\bigO(#1)}}}

\DeclareMathAlphabet{\mathsfit}{\encodingdefault}{\sfdefault}{m}{sl}
\SetMathAlphabet{\mathsfit}{bold}{\encodingdefault}{\sfdefault}{bx}{n}

\def\1{\mymacro{\bm{1}}}
\def\0{\mymacro{\bm{0}}}

\newcommand{\params}{{\mymacro{\vtheta}}}

\newcommand{\lm}{{\mymacro{p_\params}}}

\newcommand{\CODIAcr}{\mymacro{{CODI}}\xspace}
\newcommand{\PCCoTAcr}{{PCCoT}\xspace}
\newcommand{\ccntAcr}{{Coconut}\xspace}
\newcommand{\SiMCoTAcr}{{SIM-CoT}\xspace}
\newcommand{\KaVaAcr}{{KaVa}\xspace}
\newcommand{\CoTAcr}{{CoT}\xspace}

\newcommand{\LPTAcr}{\mymacro{{LPT}}\xspace}
\newcommand{\LPTsAcr}{{LPTs}\xspace}

\newcommand{\blockBudget}{\mymacro{K}}          
\newcommand{\blockidx}{\mymacro{i}}             
\newcommand{\numLoopPasses}{\mymacro{R}}        
\newcommand{\loopPassIdx}{\mymacro{t}}          
\newcommand{\numCoTSteps}{\mymacro{S}}          
\newcommand{\numCoTTokens}{\mymacro{N}}         
\newcommand{\tokPerBlock}{\mymacro{c}}          
\newcommand{\stepLoss}{\mymacro{\mathcal{L}_{\mathrm{step}}}}   
\newcommand{\ansLoss}{\mymacro{\mathcal{L}_{\mathrm{ans}}}}     
\newcommand{\totalLoss}{\mymacro{\mathcal{L}}}  
\newcommand{\loopHidden}[2][]{\mymacro{\vh^{(#2)}_{#1}}} 
\newcommand{\auxHidden}[2][]{\mymacro{\widetilde{\vz}^{(#2)}_{#1}}} 
\newcommand{\ansHidden}{\mymacro{\vz}} 
\newcommand{\latTokenSym}{\mymacro{\langle\texttt{lat}\rangle}} 
\newcommand{\latRegionStart}{\mymacro{\langle\texttt{BoT}\rangle}} 
\newcommand{\latRegionEnd}{\mymacro{\langle\texttt{EoT}\rangle}}   
\newcommand{\baseLM}{\mymacro{f_\params}}       
\newcommand{\lmhead}{\mymacro{f_{\mathrm{head}}}}   
\newcommand{\auxdec}{\mymacro{g_\phi}}              
\newcommand{\auxhead}{\mymacro{g_{\mathrm{head}}}}  
\newcommand{\auxStepLoss}{\mymacro{\mathcal{L}_{\mathrm{step}}^{\mathrm{aux}}}} 
\newcommand{\auxLoss}{\mymacro{\mathcal{L}^{\mathrm{aux}}}}                     
\newcommand{\inputSeq}{\mymacro{\boldsymbol{x}}}             
\newcommand{\loopEmb}{\mymacro{\mE}}  
\newcommand{\cache}{\mymacro{\mathcal{C}}}        
\newcommand{\embedDimLM}{\mymacro{d}}           
\newcommand{\stepTokens}[1]{\mymacro{T_{#1}}}  
\newcommand{\stepToken}[2]{\mymacro{T_{#1,#2}}} 
\newcommand{\normStepToks}{\mymacro{N_{\mathrm{step}}}} 
\newcommand{\stepLossWeight}{\mymacro{\lambda_{\mathrm{step}}}} 
\newcommand{\CE}{\mymacro{\mathrm{CE}}}         
\newcommand{\dataChainDist}{\mymacro{q}}      
\newcommand{\thoughtSeq}{\mymacro{T}}           
\newcommand{\questionVar}{\mymacro{Q}}          

\newcommand{\instr}{{\mymacro{Q}}}
\newcommand{\outstr}{{\mymacro{A}}}

\def\vh{{\mymacro{\bm{h}}}}

\def\vz{{\mymacro{\bm{z}}}}

\newcommand{\vtheta}{{\mymacro{\boldsymbol{\theta}}}}

\def\mE{{{\mymacro{\bm{E}}}}}



%% file: abstract.tex
Language models typically reason via explicit chain-of-thought (\CoTAcr), generating intermediate steps token-by-token.
Latent \CoTAcr offers an alternative: it performs multi-step reasoning in the model's hidden states, replacing decoded tokens with continuous representations for greater efficiency.
However, existing latent \CoTAcr methods underperform explicit \CoTAcr beyond $1$B parameters, and the gap widens with scale.
\emph{Looped}, or recurrent-depth, Transformers, which reuse their weights to increase computation depth without adding parameters, are a natural fit for latent reasoning.
We therefore ask whether looped Transformers can bridge this gap.
We answer affirmatively with a simple recipe: a \emph{looped padded Transformer} that processes $K$ latent blocks in parallel for $R$ iterations, with a cross-entropy loss on each latent position's gold \CoTAcr-step token, similar to explicit \CoTAcr supervision.
We instantiate it as \textbf{\modelName} (\underline{Lo}oped \underline{T}ransformers with parallel s\underline{u}pervision on latent\underline{s}).
\modelName{} is, to our knowledge, the first latent-\CoTAcr method to bridge the gap to explicit \CoTAcr at the 3B scale, while cutting thought-phase latency by $2.5\times$--$6.9\times$ from compact math expressions to natural language.
Projecting \modelName's post-loop latents through the base LM head recovers the gold reasoning steps and even surfaces alternative valid intermediate steps, evidence that its latent space is interpretable and \CoTAcr-aligned.
Ablations confirm that both the looped backbone and the parallel supervision on gold \CoTAcr tokens are essential. Code is available at \url{https://github.com/yingfan-bot/lotus}.

%% file: intro.tex
\section{Introduction}
\begin{figure}[t]
    \centering
    \includegraphics[width=\linewidth]{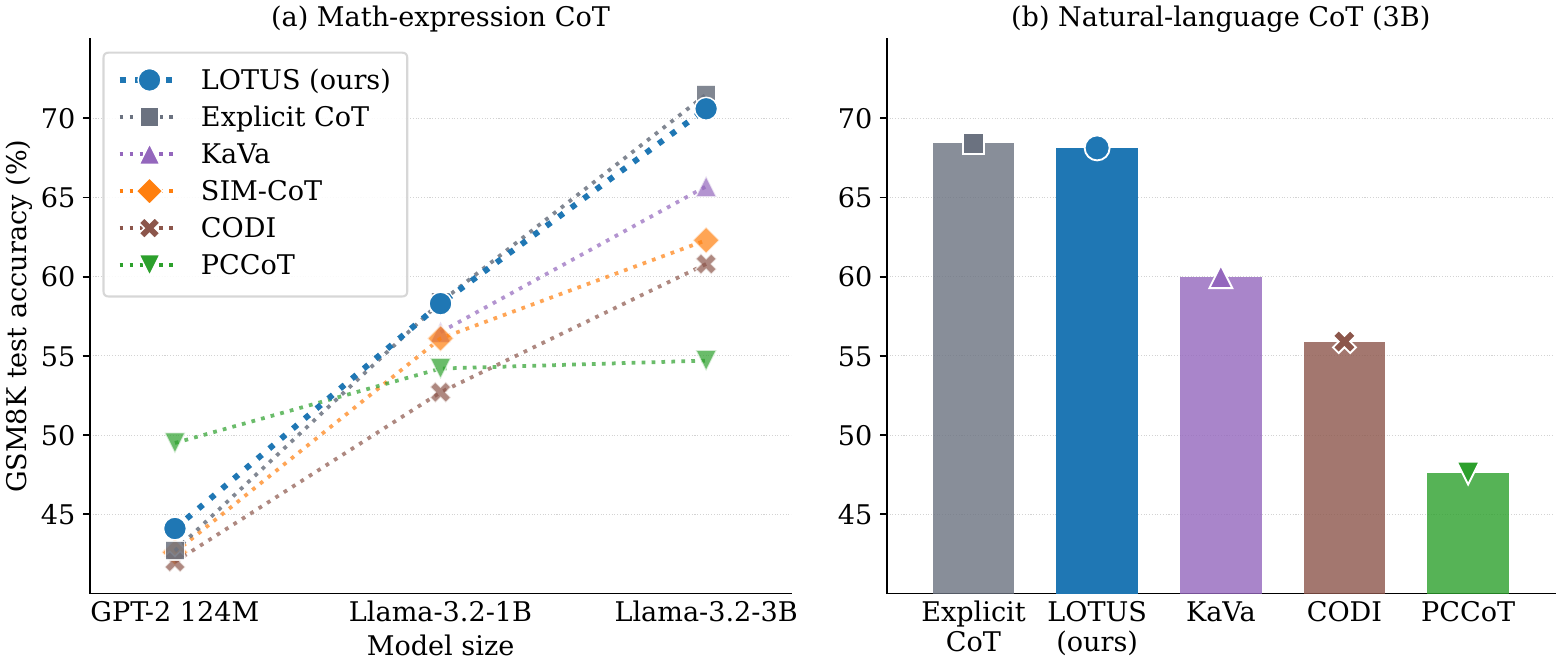}
    \caption{\modelName{} bridges the latent--explicit \CoTAcr accuracy gap across scale on GSM8K test. \textbf{(a)}~Math-expression \CoTAcr across model sizes: prior latent methods fall further below explicit \CoTAcr as the backbone grows, whereas \modelName{} tracks the explicit-\CoTAcr ceiling while cutting thought-phase latency by {$2.5\times$} at 3B. \textbf{(b)}~Natural-language \CoTAcr at 3B: \modelName{} matches explicit \CoTAcr and outperforms the latent baselines while reducing thought-phase latency by {$6.9\times$}. }
    \label{fig:teaser}
    \vspace{-2em}
\end{figure}

Scaling inference compute, i.e., letting a model ``think'' before it answers, has become a dominant lever for increasing language model capabilities, with stronger performance now coming from longer reasoning chains rather than from model size alone~\citep{deepseek-r1,openai2026openaio1card}.
Chain-of-thought (\CoTAcr) reasoning~\citep{10.5555/3600270.3602070}, where the model emits intermediate reasoning steps, drives this trend.
However, since each token must be decoded sequentially, generating a \CoTAcr of length $\numCoTTokens$ takes $\numCoTTokens$ sequential model evaluations, making reasoning costly.

\emph{Latent reasoning} aims to achieve the same at a fraction of the cost: it carries out the intermediate computation in continuous hidden states rather than decoded tokens, condensing many steps into fewer model evaluations.
On small backbones such as \textsc{GPT-2}~\citep{radford2019language}, latent methods~\citep{hao2025training,shen-etal-2025-codi,wei2025simcotsupervisedimplicitchainofthought} already match \CoTAcr accuracy.
Yet at the scales where \CoTAcr matters most, the promise breaks down: beyond $1$B parameters, no existing latent method matches explicit \CoTAcr on math reasoning, and the gap widens with model size~\citep{wei2025simcotsupervisedimplicitchainofthought}.

We identify two common issues in latent methods. 
\textbf{(P1)} Sequential generation: methods like \ccntAcr~\citep{hao2025training}, CODI~\citep{shen-etal-2025-codi}, and \SiMCoTAcr~\citep{wei2025simcotsupervisedimplicitchainofthought} produce latent thoughts in a pure autoregressive way, so the number of sequential forward passes still scales linearly with the number of latent tokens, keeping \CoTAcr's sequential generation bottleneck while giving up the readable intermediate steps that explicit \CoTAcr provides. 
\textbf{(P2)} Lack of \CoTAcr grounding: Explicit \CoTAcr gives every reasoning step a direct, position-aligned token target. 
Without similarly grounded supervision, latent traces can drift from meaningful computation and destabilize training at scale~\citep{li2026chainthoughtcompressiontheoritical,zou2026capabilitiesfundamentallimitslatent}. 
Methods like \PCCoTAcr~\citep{wu-etal-2025-parallel} and \KaVaAcr~\citep{kuzina2026kava} mitigate the sequential bottleneck with Jacobi-style iteration but still ground their latents through indirect signals that fall short of \CoTAcr-aligned target, such as hidden-state distillation from a teacher \CoTAcr model or a compressed teacher key--value cache.

An architecture that resolves both issues at once would refine its latents in a few parallel passes rather than one autoregressive step per token, and would ground them in a target as direct as explicit \CoTAcr's.
\emph{Looped}, or recurrent-depth, Transformers are a natural fit for the first requirement: they reuse the same weights across iterations to add computation depth without extra parameters, a design that recent work scales to billion-parameter pretraining~\citep{geiping2025scalingup,zhu2025scalinglatentreasoninglooped,zeng2026ponderlm}.
On parallelizable problems, they can provably reach a solution in fewer model iterations \CoTAcr~\citep{saunshi2025reasoninglatentthoughtspower}; for example, \LPTsAcr solve graph reachability in a logarithmic number of iterations~\citep{merrill2025littledepthgoeslong}---an exponential improvement over sequential latent methods.
This raises the question we study: \emph{can a looped Transformer bridge the gap to explicit \CoTAcr while reasoning in fewer sequential steps?}

We find that pairing a looped padded backbone with simple supervision on the gold \CoTAcr tokens is surprisingly effective. We present \textbf{\modelName} (\underline{Lo}oped \underline{T}ransformers with parallel s\underline{u}pervision on latent\underline{s}).
\modelName{} places $\blockBudget$ learnable padded latent blocks of $\tokPerBlock$ tokens each between the question and answer, and processes the sequence with the base model $\numLoopPasses$ times. An explicit \CoTAcr of length $\numCoTTokens$ is thus processed in $\numLoopPasses \!\ll\! \numCoTTokens$ iterations of dense parallel computation rather than $\numCoTTokens$ sequential per-token generations, which addresses \textbf{P1}. 
To address \textbf{P2}, \modelName{} supervises the latents directly through the base model's own LM head: a cross-entropy loss aligns each latent position to its corresponding gold \CoTAcr-step token, all in parallel. The supervision target can also be routed through an auxiliary decoder that is conditioned on all latents and scores the gold \CoTAcr tokens under teacher forcing, which we instantiate on the identical looped backbone as \modelName-aux. Both routings reach near-\CoTAcr accuracy at the 3B scale.
\modelName{} is, to our knowledge, the first latent reasoning method at the \textsc{Llama-3.2-3B-Instruct} scale to bridge the in-domain gap to explicit \CoTAcr on GSM8k~{(\cref{fig:teaser})}, while surpassing \CoTAcr on the out-of-domain average and cutting thinking time by $\mathbf{2.5\times}$. On a more verbose natural-language \CoTAcr stress test, \modelName{} is on par with explicit \CoTAcr in accuracy while reducing thought-phase latency by $\mathbf{6.9\times}$.
\modelName's latent space is also interpretable: reading the post-loop latents through the base LM head recovers the gold reasoning steps, and even surfaces alternative valid intermediate steps the model was never trained on, evidence that the latents are genuinely \CoTAcr-aligned rather than opaque.
We summarize our contributions as follows:
\begin{itemize}[leftmargin=*, noitemsep, topsep=0pt]
    \item We propose the recipe of looped padded Transformers with parallel cross-entropy supervision on gold \CoTAcr tokens. The method uses a padded latent prefix (\cref{sec:method:prefix}), refines it with a looped backbone (\cref{sec:method:loop}), and supervises the post-loop latents through the base LM head (\cref{sec:method:supervision}). We also instantiate \modelName-aux, which routes the same latent supervision through an auxiliary decoder used only during training (\cref{sec:method:lotus-aux}).
    \item We show that \modelName{} is, to our knowledge, the first latent reasoning method to bridge the latent--explicit \CoTAcr gap on math reasoning at the 3B scale: on \textsc{Llama-3.2-3B-Instruct}, it brings the GSM8K test accuracy to within $1$ point of explicit \CoTAcr and surpasses explicit \CoTAcr on the out-of-domain math average. It also cuts thought-phase latency by $2.5\times$ in the math-expression setting and by $6.9\times$ in the natural-language \CoTAcr stress test (\cref{sec:exp:main}).
    \item Ablations show that both the parallel supervision design (\cref{sec:exp:ablation:supervision}) and the looped architecture design (with enough block width and loop depth, \cref{sec:exp:ablation:loop}) are important. The direct LM-head routing is robust across scale, whereas the auxiliary decoder routing matches it at 3B but degrades on smaller backbones (\cref{sec:exp:main}).
    \item \modelName{} yields a transparent, \CoTAcr-aligned latent space. Reading the post-loop latents recovers most of the gold intermediate tokens (\cref{sec:exp:analysis:cross-reader}) while assigning nontrivial mass to unseen-but-valid alternatives (\cref{sec:exp:analysis:multipath}). A loss ablation shows that the \CoTAcr supervision anchors the readout to the gold chain, while answer supervision helps select the coherent joint (\cref{sec:exp:analysis:loss-ablation}).
\end{itemize}

%% file: prelim.tex
\section{Preliminaries}
\label{sec:prelim}
A glossary of all notation is provided in \cref{tab:notation} in \cref{app:notation}.
\paragraph{Explicit CoT.}
A \defn{language model} (LM) $\baseLM$ can map a question $\instr$ to an answer $\outstr$ via multi-step reasoning: generating $\numCoTSteps$ intermediate steps $\stepTokens{1}, \ldots, \stepTokens{\numCoTSteps}$ before the answer, where $\stepTokens{\blockidx}$ ($\blockidx\in\{1,\dots,\numCoTSteps\}$) may span multiple tokens. We write $\numCoTTokens$ for the total number of CoT tokens across all $\numCoTSteps$ steps.
Standard \CoTAcr supervision trains all CoT tokens in parallel via teacher forcing, but at inference time the tokens must be generated sequentially, increasing latency. 
We refer to this setup as \emph{Explicit} CoT.

\paragraph{Latent reasoning} replaces decoded CoT tokens with multi-step inference in the model's hidden states, which are commonly referred to as latent thoughts. 
\citet{zhu2025reasoning} show latent thoughts carry strictly more information than discrete tokens: linearly many latent steps in a Transformer can solve graph reachability that would otherwise require quadratically many explicit \CoTAcr steps.

\paragraph{Looped padded Transformers.}
A \defn{looped Transformer}~\citep{dehghani2019universaltransformers} applies $\baseLM$'s backbone multiple times, gaining depth without adding parameters. \emph{Padding} the input with extra learnable tokens~\citep{goyal2024think,pfau2024lets} adds parallel computation to each iteration. The combination of the two creates \defn{looped padded Transformers} (\LPTsAcr), which are more capable than either component alone.

\paragraph{Prior latent reasoning methods.}
We contrast \modelName{} against the most closely related latent reasoning methods, which we summarize here (details in \cref{app:differentiation}).
\emph{Coconut}~\citep{hao2025training} replaces each \CoTAcr step with a $\tokPerBlock$ latent tokens generated autoregressively from the previous one, under a curriculum, and supervises only the answer.
CODI~\citep{shen-etal-2025-codi} keeps this autoregressive latent decoding but adds a distillation target, aligning one designated latent's hidden state with the corresponding hidden state of a teacher \CoTAcr model.
SIM-CoT~\citep{wei2025simcotsupervisedimplicitchainofthought} instead trains an auxiliary autoregressive decoder that aligns each latent with its \CoTAcr token.
\PCCoTAcr~\citep{wu-etal-2025-parallel} and \KaVaAcr~\citep{kuzina2026kava} drop the autoregressive nature, refining a fixed budget of latent tokens in parallel over a few Jacobi iterations; they ground the latents indirectly, through CODI-style distillation and a compressed teacher key--value cache, respectively.
We refer the reader to \cref{app:related} for an extended discussion of related work on latent reasoning and looped Transformers, and to \cref{app:differentiation} for a detailed comparison with the related latent reasoning methods.

%% file: method.tex
\section{Method}
\label{sec:method}

\begin{figure}[t]
\includegraphics[width=1.0\linewidth]{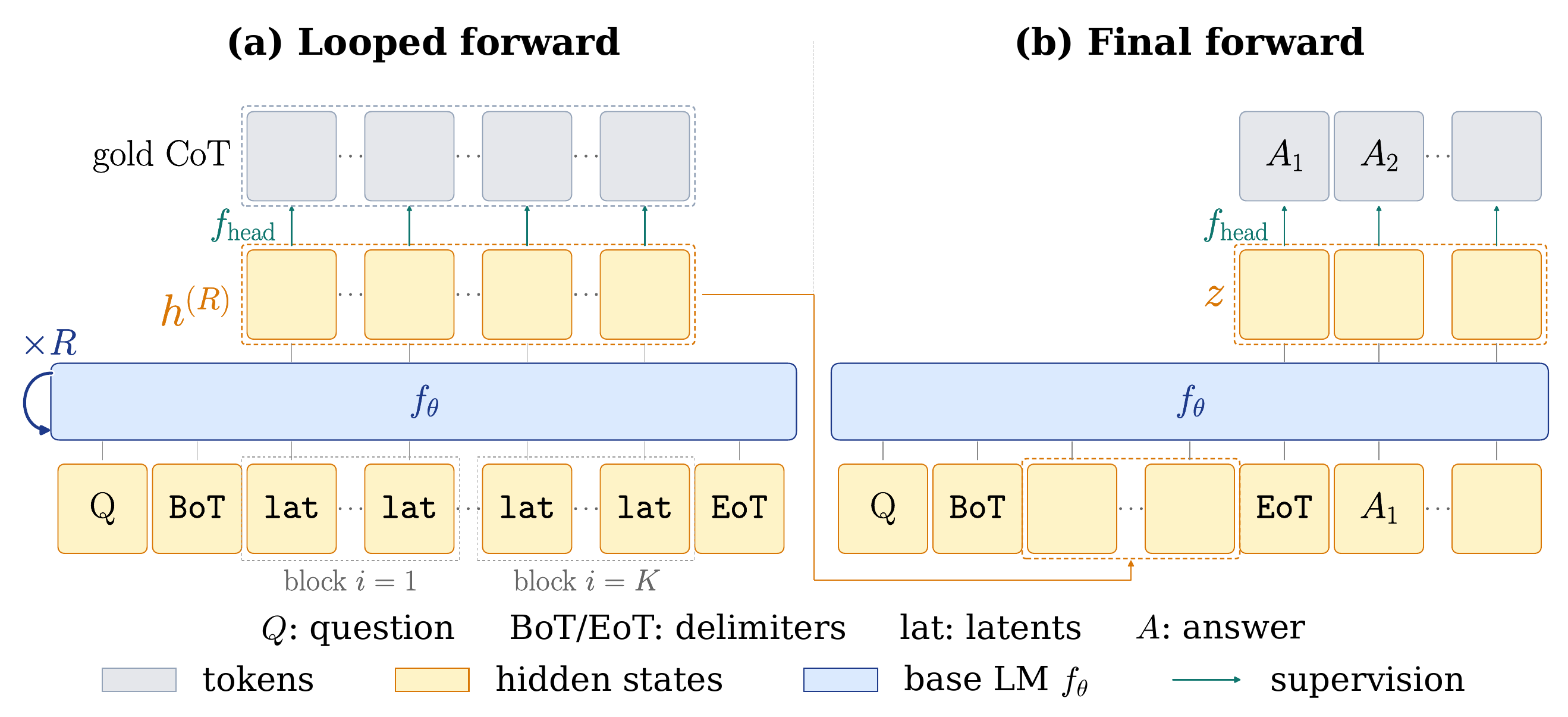}
    \caption{\modelName architecture. (a) Looped forward: the looped LM $\baseLM$ is iterated $\numLoopPasses$ times over $[\latRegionStart, \latTokenSym\!\cdots\!\latTokenSym, \latRegionEnd]$, producing post-loop hidden states $\loopHidden{\numLoopPasses}$ at the latent positions, and $\stepLoss$ supervises these states through the LM head, in parallel, against the corresponding \CoTAcr tokens $T_{i,j}$. (b) Final forward: the post-loop latents $\loopHidden{\numLoopPasses}$ are inserted at the latent positions and the answer suffix $\outstr$ is supervised by next-token prediction ($\ansLoss$) through the LM head.}
    \label{fig:dslt}
\end{figure}

Although \LPTsAcr{} are expressive enough to carry out efficient parallel reasoning, without structured supervision they often fail to generalize out-of-distribution~\citep{altabaa2025unlockingoutofdistributiongeneralizationtransformers}.
This motivates our central design choice: grounding the latent computation directly in the gold \CoTAcr tokens.
We turn a standard LM into a latent reasoner with a recipe built from two ingredients:
\begin{enumerate*}[label=\textit{(\roman*)}]
\item a \emph{padded latent prefix processed by an \LPTAcr}, and
\item \emph{parallel cross-entropy supervision on exact gold \CoTAcr tokens}.
\end{enumerate*}
We instantiate this recipe as \textbf{\modelName} (\underline{Lo}oped \underline{T}ransformers with parallel s\underline{u}pervision on latent\underline{s}, \cref{fig:dslt}), which reads the supervision directly through the base model's LM head. We also study \modelName-aux (\cref{fig:dslt-aux}), which instead routes the supervision through an auxiliary decoder.
We detail the padded latent prefix in \cref{sec:method:prefix}, the looped computation in \cref{sec:method:loop}, and \modelName's objective---together with an analysis of why it works (\cref{sec:method:theory})---in \cref{sec:method:supervision}, then introduce the \modelName-aux variant in \cref{sec:method:lotus-aux}.

\subsection{Padded Latent Prefix}
\label{sec:method:prefix} 

For a question $\instr$, $\numCoTSteps$ chain-of-thought (CoT) steps $\stepTokens{1},\dots,\stepTokens{\numCoTSteps}$, and answer $\outstr = (A_1, \dots, A_{|\outstr|})$ (with $A_{|\outstr|} = \langle\text{EoS}\rangle$), we construct an input
\begin{equation}
    \inputSeq \;=\; \big[\,\instr,\; \latRegionStart,\;
    \underbrace{\latTokenSym\cdots\latTokenSym}_{\text{block } 1\;(\tokPerBlock \text{ tokens})},\;
    \cdots,\;
    \underbrace{\latTokenSym\cdots\latTokenSym}_{\text{block } \blockBudget\;(\tokPerBlock \text{ tokens})},\;
    \latRegionEnd,\; \outstr\,\big],
    \label{eq:prefix}
\end{equation}
where $\latTokenSym$ is a learnable latent token shared across positions, and $\latRegionStart, \latRegionEnd$ are learnable special tokens that delimit the latent region. The block budget $\blockBudget$ and per-block width $\tokPerBlock$ are fixed hyperparameters, so the latent region contains $\blockBudget\tokPerBlock$ latent tokens plus the two delimiters. The latent prefix follows prior work on padding or pause tokens~\citep{goyal2024think,pfau2024lets}, with each block $\blockidx$ intended to encode the latent representation of \CoTAcr step $\stepTokens{\blockidx}$, $\blockidx\in\{1,\dots,\blockBudget\}$.
Unlike the example-dependent step count $\numCoTSteps$ of \cref{sec:prelim}, the block budget $\blockBudget$ is fixed across examples; we choose it to cover the step counts in our data, so that $\blockBudget\ge\numCoTSteps$ on almost all examples and each block can align to one \CoTAcr step (\cref{app:details}).

\subsection{Looped Latent Computation}
\label{sec:method:loop}

Let $\baseLM$ be the base causal LM and $\loopEmb\in\R^{\blockBudget\tokPerBlock \times \embedDimLM}$ the learnable latent embeddings, indexed as $\loopEmb_{\blockidx,j}$ for block $\blockidx\in\{1,\dots,\blockBudget\}$ and intra-block position $j\in\{1,\dots,\tokPerBlock\}$.
\modelName{} first runs a single forward pass over the prefix $[\instr, \latRegionStart]$ to populate a KV cache $\cache_\text{pre}$, which is reused throughout training and inference without recomputation. It then iterates $\baseLM$ for $\numLoopPasses$ iterations, each refining the latent embeddings while attending to $\cache_\text{pre}$ (the trailing $\latRegionEnd$ and the answer suffix $\outstr$ do not enter the loop). Writing $\loopHidden{\loopPassIdx}\in\R^{\blockBudget\tokPerBlock \times \embedDimLM}$ for the hidden states at the latent positions after iteration $\loopPassIdx$, with entries $\loopHidden[\blockidx,j]{\loopPassIdx}$:
\begin{equation}
\begin{aligned}
    \loopHidden{0} \;&=\; \baseLM\!\left( \loopEmb \;\middle|\; \cache_\text{pre}\right), \\
    \loopHidden{\loopPassIdx} \;&=\; \baseLM\!\left( \loopEmb + \loopHidden{\loopPassIdx-1}\;\middle|\; \cache_\text{pre}\right), \quad \loopPassIdx = 1,\dots,\numLoopPasses.
\end{aligned}
\end{equation}
where the brackets denote subsequence concatenation. This is a finite-unroll, input-injected recurrence over a looped
Transformer~\citep{dehghani2019universaltransformers,fan2025looped},
with a fixed unroll depth $\numLoopPasses$ and gradients
propagated through all $\numLoopPasses$ iterations.

\subsection{Direct Step-Aligned Supervision}
\label{sec:method:supervision}

\subsubsection{Objective}
\label{sec:method:supervision:objective}

\paragraph{Step \CoTAcr supervision loss.}
After $\numLoopPasses$ iterations of looping (\cref{sec:method:loop}), we supervise each latent position $(\blockidx,j)$ in the grid with the corresponding \CoTAcr step token. Let $\stepTokens{\blockidx} = (\stepToken{\blockidx}{1},\dots,\stepToken{\blockidx}{\tokPerBlock})$ be the tokenized \CoTAcr step $\blockidx$ padded or truncated to $\tokPerBlock$ tokens. We apply a single batched cross-entropy that directly aligns each block to its target step:
\begin{equation}
    \stepLoss
    \;=\; \frac{1}{\normStepToks}\sum_{\blockidx=1}^{\blockBudget}\sum_{j=1}^{\tokPerBlock}
    \CE\!\left(\lmhead(\loopHidden[\blockidx,j]{\numLoopPasses}),\, \stepToken{\blockidx}{j}\right),
    \label{eq:step}
\end{equation}
where $\lmhead$ is the LM head and $\normStepToks = \sum_{\blockidx} |\stepTokens{\blockidx}|$ is the total number of supervised (non-padding) \CoTAcr tokens.\footnote{Padding tokens used to pad CoT steps to length $\tokPerBlock$ are ignored by the cross-entropy (and excluded from $\normStepToks$).}

\paragraph{Answer supervision loss.}
$\ansLoss$ is computed in a separate \emph{final forward} of $\baseLM$ that reuses the prefix cache $\cache_\text{pre}$ and inserts the post-loop latent hidden states $\loopHidden{\numLoopPasses}$ at the latent rows:
\begin{equation}
    \ansHidden \;=\; \baseLM\!\left( [\latRegionEnd,\, \outstr] \;\middle|\; [\cache_\text{pre},\, \loopHidden{\numLoopPasses}]\right),
    \label{eq:final-forward}
\end{equation}
where $\ansHidden_m\in\R^{\embedDimLM}$
collects the resulting hidden state at each answer-suffix position $m\in\{0, 1, \dots, |\outstr|-1\}$ ($m=0$ being the trailing $\latRegionEnd$ position, which predicts $A_1$). We denote hidden states here by $\ansHidden$, distinct from the looped forward latents $\loopHidden{\loopPassIdx}$, to mark that they come from different sources. Applying the LM head $\lmhead$, standard next-token cross-entropy on the answer suffix gives:
\begin{equation}
    \ansLoss \;=\; \frac{1}{|\outstr|}\sum_{m=0}^{|\outstr|-1}
    \CE\!\left(\lmhead(\ansHidden_m),\, A_{m+1}\right).
\end{equation}
Given a step supervision weight $\stepLossWeight$, the full objective is
\begin{equation}
    \totalLoss \;=\; \ansLoss + \stepLossWeight\,\stepLoss.
    \label{eq:loss}
\end{equation}
 
\paragraph{Key properties of the objective.}
The supervision in \cref{eq:loss} has three properties: 
\begin{enumerate*}[label=\textit{(\roman*)}]
    \item it is \emph{direct}: the per-block target $\stepTokens{\blockidx}$ is scored through the same LM head used to produce the answer;
    \item it is \emph{parallel}: all $\blockBudget$ blocks are supervised simultaneously; and
    \item it is \emph{post-loop}: the readout reads $\vh^{(\numLoopPasses)}$ after the final iteration, rather than at every iteration.
\end{enumerate*}
We ablate the design choice in \cref{sec:exp:ablation}, and compare against prior latent reasoning methods in \cref{app:differentiation}.

\paragraph{Inference.} At inference, \modelName{} runs the looped forward and then decodes the answer. We compute the KV cache of the question $\instr$ once, iterate the loop $\numLoopPasses$ times to obtain the post-loop latents $\loopHidden{\numLoopPasses}$, and then decode the answer $\outstr$ autoregressively through the base LM head while conditioning on those latents (same as \cref{fig:dslt} only without computing the losses). The reasoning is carried entirely by the parallel latent blocks, so the only sequential decoding is over the short answer suffix, which is the source of the latency gains reported in \cref{sec:exp:main}. 

\paragraph{No latent decoding for answer generation.} Note that the latent reasoning steps are never read out during answer generation. They are projected to the token space through the LM head only to align them with CoT reasoning during training with $\stepLoss$.
The answer is always generated from the latent hidden states, both in the final-forward $\ansLoss$ pass (\cref{eq:final-forward}) and at inference.

\subsubsection{Roles of \texorpdfstring{$\stepLoss$ and $\ansLoss$}{the step and answer losses}}
\label{sec:method:theory}

A natural worry about the \modelName{} objective (\cref{eq:loss}) is that it supervises each latent position \emph{independently}: \cref{eq:step} scores every gold \CoTAcr token in parallel, with no autoregressive factorization of the chain.
Why should independent per-position targets train a model that produces a globally coherent answer?
We give a lens that resolves this, distinguishing the role of the two losses and explaining why both are needed.

\paragraph{Parallel chain likelihood.}
The step loss $\stepLoss$ supervises a \emph{parallel chain likelihood} (PCL) over the readout positions.
Maximizing the per-position probability $p_\params(\stepToken{\blockidx}{j} \mid \questionVar)$ of each gold token at its latent position induces the chain likelihood
\begin{equation}
p_\params^\mathrm{PCL}(\thoughtSeq\mid\questionVar)
=\prod_{\blockidx=1}^{\blockBudget}\prod_{j=1}^{\tokPerBlock}
p_\params(\stepToken{\blockidx}{j}\mid\questionVar),
\label{eq:pcl}
\end{equation}
rather than the autoregressive conditionals $p_\params(\stepToken{\blockidx}{j} \mid \questionVar, \thoughtSeq_{<\blockidx}, \stepToken{\blockidx}{<j})$.
The factorization treats the readout as conditionally independent across positions, but the latent states themselves are not independent: the looped Transformer computes them \emph{jointly} over the padded workspace, so the dependence the factorization drops is carried by the shared latent computation rather than by the loss.

\paragraph{Complementary roles.} This view makes the division of labor between the two losses precise. First, the support inclusion
\begin{equation}
    \mathrm{supp}\!\left(\dataChainDist(\thoughtSeq \mid \questionVar)\right) \;\subseteq\; \prod_{\blockidx=1}^{\blockBudget}\prod_{j=1}^{\tokPerBlock} \mathrm{supp}\!\left(\dataChainDist(\stepToken{\blockidx}{j} \mid \questionVar)\right)
    \label{eq:support}
\end{equation}
shows that every gold chain lies inside the Cartesian product of per-position gold token supports.\footnote{Minimizing $\stepLoss$, the cross-entropy to these marginals, pulls the model's PCL toward that product, i.e.\ minimizes $\mathrm{KL}\big(\prod_{\blockidx,j}\dataChainDist(\stepToken{\blockidx}{j}\mid\questionVar)\,\big\|\,p_\params^\mathrm{PCL}(\thoughtSeq\mid\questionVar)\big)$.}
So $\stepLoss$ plays a \emph{support-coverage} role: it makes each position place mass on the right gold tokens, without needing to reproduce the joint distribution---which \modelName{} never samples, since the answer is decoded directly from the jointly computed latent states.
Second, $\ansLoss$ supplies the global \emph{selection} pressure that PCL alone lacks: because the answer is trained while conditioning on the entire latent configuration, gradients favor jointly computed hidden states that can actually support the correct answer.
The two losses are thus complementary by construction---coverage from $\stepLoss$, joint selection from $\ansLoss$---which is exactly the behavior we verify empirically in \cref{sec:exp:analysis:loss-ablation}, where dropping either loss degrades the gold chain likelihood.

To further probe how \modelName{} compares to modeling the autoregressive chain likelihood, we introduce an autoregressive decoder variant in \cref{sec:method:lotus-aux} and compare the results in \cref{sec:exp:main}.

\subsection{\modelName-aux: An Auxiliary Decoder Variant}
\label{sec:method:lotus-aux}

An alternative route for \CoTAcr supervision explored in existing work is autoregressive chain likelihood: an auxiliary autoregressive decoder conditioned on the latent blocks scores the \CoTAcr tokens
under teacher forcing. This creates a trade-off: in exchange for modeling the chain autoregressively, \CoTAcr supervision no longer flows directly through the answer-generating base LM head, and teacher forcing introduces an extra train/inference mismatch.

We instantiate this routing on our looped padded backbone as \modelName-aux. \modelName-aux reuses the same looped forward and final forward as \modelName{} (\cref{fig:dslt}). The only difference is the supervision in the looped forward, which is routed through an auxiliary decoder rather than read directly through the base LM head. The auxiliary decoder mirrors \SiMCoTAcr~\citep{wei2025simcotsupervisedimplicitchainofthought}, but instead of generating latents autoregressively and supervising a single latent per \CoTAcr step, \modelName-aux produces all $\blockBudget$ latent blocks in parallel through the looped backbone and supervises a full $\tokPerBlock$-token block per step. We describe this routing below.

\begin{wrapfigure}[18]{r}{0.4\linewidth}
    \centering
    \vspace{-1.2em}
    \includegraphics[width=\linewidth]{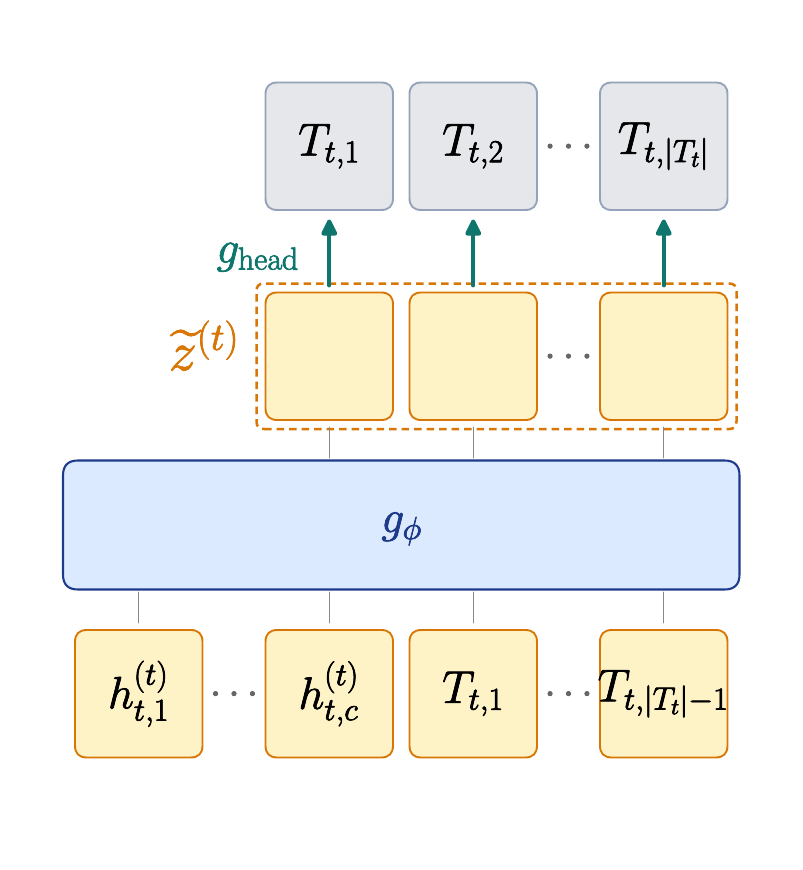}
    \caption{\modelName-aux supervision at loop iteration $t$ for block $t$. The auxiliary decoder $\auxdec$ only replaces the base LM head supervision in \cref{fig:dslt}(a).}
    \label{fig:dslt-aux}
    \vspace{-1.5em}
\end{wrapfigure}

\paragraph{Latent supervision via the auxiliary decoder.}
As shown in Figure~\ref{fig:dslt-aux}, the auxiliary decoder $\auxdec$ is an extra decoder model that, at each loop iteration $t$, reads latent block $\loopHidden{t}_t$ (we select block $i=t$ at iteration $t$, the $\tokPerBlock$ post-loop latents of the $t$-th block) as a $\tokPerBlock$-token prefix and predicts the gold \CoTAcr step $\stepTokens{t}=(T_{t,1},\dots,T_{t,\lvert\stepTokens{t}\rvert})$ under teacher forcing, where $\lvert\stepTokens{t}\rvert$ is the step's token length. Let $\auxHidden{t}=(\auxHidden[0]{t},\dots,\auxHidden[\lvert\stepTokens{t}\rvert-1]{t})$ denote the output hidden states at each input \CoTAcr position $m\in\{0,1,\dots,\lvert\stepTokens{t}\rvert-1\}$ ($m{=}0$ being the last latent position $\loopHidden[t,\tokPerBlock]{t}$, which predicts $T_{t,1}$, and $m{>}0$ the gold token $T_{t,m}$). Mirroring the final-forward pass of \cref{eq:final-forward}, we generate $\auxHidden{t}$ with
\begin{equation}
\label{eq:aux-h}
    \auxHidden{t} \;=\; \auxdec\!\left( [\,\loopHidden[t,\tokPerBlock]{t},\; T_{t,1}, \dots, T_{t,\lvert\stepTokens{t}\rvert-1}\,] \;\middle|\; \loopHidden[t,1]{t}, \dots, \loopHidden[t,\tokPerBlock-1]{t} \right),
\end{equation}
where the first $\tokPerBlock-1$ latent positions $\loopHidden[t,1]{t},\dots,\loopHidden[t,\tokPerBlock-1]{t}$ are the conditioning prefix. The teacher-forced next-token cross-entropy on the gold \CoTAcr gives:
\begin{equation}
\label{eq:aux}
    \auxStepLoss \;=\; \frac{1}{N_\text{step}} \sum_{t=1}^{\blockBudget} \sum_{m=0}^{\lvert\stepTokens{t}\rvert-1} \CE\!\left(\auxhead(\auxHidden[m]{t}),\; T_{t,m+1}\right),
\end{equation}
where $N_\text{step}=\sum_{t=1}^{\blockBudget}\lvert\stepTokens{t}\rvert$ is the total number of supervised \CoTAcr tokens. The supervision itself remains parallel: under teacher forcing the gold prefix is fed at every position, so all \CoTAcr positions are scored in a single forward pass with no sequential generation.
In other words, the autoregressive factorization affects only the loss, not the computation.

\paragraph{Per-iteration supervision and the full objective.}
Equation~\ref{eq:aux} reads block $i=t$ at iteration $t$ (per-iteration, rather than at the final iteration $R$). We validate this choice in \cref{sec:exp:ablation:supervision}, where per-iteration readouts outperform reading all blocks at the final iteration $\numLoopPasses$ for the auxiliary decoder routing. The answer loss $\ansLoss$ from \cref{sec:method:supervision} is unchanged, so the full \modelName-aux objective is $\auxLoss = \ansLoss + \stepLossWeight\,\auxStepLoss$. In our experiments $\auxdec$ is initialized with a copy of the base LM and trained together with the base LM. Training and architectural details are in \cref{app:details}.

\paragraph{Inference.}
The auxiliary decoder $\auxdec$ is used only at training time. At inference, \modelName-aux generates the final answer autoregressively from the post-loop latents exactly as \modelName{} does (\cref{fig:dslt}(b)), so the inference efficiency gain of \modelName{} also applies to \modelName{}-aux.

%% file: results.tex

\section{Experiments}
\label{sec:exp}

Our experiments investigate four questions: how accurate \modelName{} is, how efficient it is at inference, how much each design choice contributes, and whether its latent space is interpretable and \CoTAcr-aligned.
\Cref{sec:exp:setup} introduces the experimental setup, \cref{sec:exp:main} presents accuracy and latency results against explicit \CoTAcr and prior latent baselines, \cref{sec:exp:ablation} ablates the supervision design, latent block width $\tokPerBlock$, loop depth $\numLoopPasses$, and inference-time robustness to changing $\tokPerBlock$ and $\numLoopPasses$ without retraining, and \cref{sec:exp:analysis} analyzes the learned latent representations.
Full training and evaluation details are in \cref{app:details}. 

\subsection{Setup}
\label{sec:exp:setup}

\paragraph{Datasets and backbones.}
Following SIM-CoT~\citep{wei2025simcotsupervisedimplicitchainofthought}, we train on GSM8k-Aug~\citep{deng2023implicit} with 385k training samples and report in-domain accuracy on the GSM8K test set~\citep{cobbe2021trainingverifierssolvemath}, together with three out-of-domain benchmarks: GSM-Hard, MultiArith, and SVAMP. We additionally train on a natural-language version of GSM8K-Aug~\citep{wu-etal-2025-parallel} as an efficiency stress test, where each reasoning step is a full-sentence rationale rather than a compact math expression. We use three backbones in increasing size: \textsc{GPT-2}~(124M)~\citep{radford2019language}, \textsc{Llama-3.2-1B-Instruct}, and \textsc{Llama-3.2-3B-Instruct}~\citep{grattafiori2024llama}.

\paragraph{Methods.}
We report \textbf{\modelName} ($\blockBudget{=}6$ blocks of $\tokPerBlock{=}25$ tokens on the \textsc{Llama} backbones, $\tokPerBlock{=}13$ on \textsc{GPT-2}, and $\numLoopPasses{=}6$ looped iterations)\footnote{$\blockBudget$ is chosen to cover the target maximum CoT step count: GSM8K solutions have at most six steps for $99\%$ of examples, so $\blockBudget{=}6$ assigns one block per step and falls back to autoregressive completion otherwise.} and \textbf{\modelName + CODI}, which adds CODI's single-position distillation loss. The auxiliary decoder variants \modelName-aux{} and \modelName-aux + CODI (\cref{sec:method:lotus-aux}) share the same configuration. We compare with \textbf{Explicit \CoTAcr} and \textbf{No-CoT} as reference, and with \textbf{CODI}~\citep{shen-etal-2025-codi} and \textbf{CODI + SIM-CoT}~\citep{wei2025simcotsupervisedimplicitchainofthought} for each backbone, which share the sequential latent reasoning budget ($\numLoopPasses{=}6$) with \modelName{} and differ only in the number of latent tokens per step. Latent methods that use a different sequential compute budget: \ccntAcr~\citep{hao2025training}, \ccntAcr + SIM-CoT (with 10 sequentially generated latent thought in total), and the parallel-latent methods \PCCoTAcr~\citep{wu-etal-2025-parallel} and \KaVaAcr~\citep{kuzina2026kava} (with 3 sequential steps and 24 latent tokens in total) are compared separately in \cref{tab:pccot-kava}  and \cref{tab:nl-cot-kava-baselines}~(\cref{app:differentiation}).

\subsection{Main Results}
\label{sec:exp:main}

\paragraph{\modelName{} matches explicit \CoTAcr at scale, where prior latent methods fall behind.}
\modelName scales well from \textsc{GPT-2} to \textsc{Llama-3.2-3B-Instruct} (\cref{tab:accuracy}). Prior latent methods match explicit \CoTAcr at small scale but fall behind as the backbone grows: CODI~+~SIM-CoT is level with explicit \CoTAcr at \textsc{GPT-2} ($42.6$ vs.\ $42.7$) yet trails it by $9.2$ points at 3B, and \KaVaAcr~\citep{kuzina2026kava} widens similarly ($1.9$ to $5.8$ points, \cref{tab:pccot-kava}). \modelName{} instead stays within ${\sim}1.5$ points of explicit \CoTAcr in-domain at each scale, and its gap does not widen. While the smaller-scale results are rather clustered, the decisive comparison is at \textsc{Llama-3.2-3B-Instruct}: \modelName{} surpasses explicit \CoTAcr on the out-of-domain average and brings the in-domain GSM8K gap to within $1.5$ points. \modelName + CODI further tightens it to within $1$ point. 

\paragraph{The two routings match at 3B, but only direct \modelName{} stays robust at smaller scales.}
At 3B, \modelName-aux performs comparably to \modelName{} and well above the CODI~+~SIM-CoT baseline, which uses an auxiliary decoder of the same size as the base LM. The gains therefore come from the looped padded backbone with parallel \CoTAcr supervision, regardless of the routing. This parity holds only at 3B: at smaller scales \modelName-aux degrades, whereas direct LM-head supervision (\modelName) stays robust across scale.

\begin{table}[t]
    \centering
    \small
    \caption{Main results across three backbones (\textsc{GPT-2},
    \textsc{Llama-3.2-1B-Instruct}, \textsc{Llama-3.2-3B-Instruct}).
    Accuracy (\%) on in-domain (GSM8K) and out-of-domain
    (GSM-Hard, MultiArith, SVAMP) benchmarks. All methods are trained on
    GSM8k-Aug~\citep{deng2023implicit}.
    ``Average'' is the mean
    over the three out-of-domain datasets. Results are reported as
    mean~$\pm$~std across $3$ seeds. The 3B block is the main scaling comparison: \modelName{} nearly closes the in-domain gap to the explicit \CoTAcr reference while leading the out-of-domain average.} 
    \label{tab:accuracy}
    \setlength{\tabcolsep}{4pt}
    \begin{tabular}{l l c c c c c}
        \toprule
        & & \textbf{In-domain} & \multicolumn{4}{c}{\textbf{Out-of-domain}} \\
        \cmidrule(lr){3-3} \cmidrule(lr){4-7}
        Backbone & Method
            & GSM8K $\uparrow$
            & GSM-Hard $\uparrow$
            & MultiArith $\uparrow$
            & SVAMP $\uparrow$
            & Average $\uparrow$ \\
        \midrule
        \multirow{8}{*}{\textsc{GPT-2}}
          & Explicit \CoTAcr                 & $42.7$ & $9.0$ & $85.0$ & $41.6$ & $45.2$ \\
          & No-CoT                 & $19.1$ & $4.3$ & $41.1$ & $16.4$ & $20.6$ \\
        \cmidrule(l){2-7}
          & CODI      & $42.0$ & $9.4$ & $\mathbf{93.0}$ & $41.7$ & $\underline{48.0}$ \\
          & CODI + SIM-CoT
                                   & $42.6$ & $9.4$ & $\underline{92.8}$ & $\mathbf{42.6}$ & $\mathbf{48.3}$ \\
          & \textbf{\modelName}
            & $\mathbf{44.1{\scriptstyle\,\pm\,0.7}}$
            & $\underline{9.5{\scriptstyle\,\pm\,0.2}}$
            & $92.4{\scriptstyle\,\pm\,1.4}$
            & $\underline{41.8{\scriptstyle\,\pm\,0.9}}$
            & $47.9$ \\
          & \textbf{\modelName + CODI}
            & $\underline{43.1{\scriptstyle\,\pm\,1.2}}$
            & $\mathbf{9.8{\scriptstyle\,\pm\,0.5}}$
            & $90.0{\scriptstyle\,\pm\,4.3}$
            & $41.6{\scriptstyle\,\pm\,0.4}$
            & $47.1$ \\
          & \textbf{\modelName-aux}
            & $35.5{\scriptstyle\,\pm\,4.5}$
            & $8.2{\scriptstyle\,\pm\,1.2}$
            & $82.0{\scriptstyle\,\pm\,2.8}$
            & $35.5{\scriptstyle\,\pm\,1.5}$
            & $41.9$ \\
          & \textbf{\modelName-aux + CODI}
            & $38.1{\scriptstyle\,\pm\,3.7}$
            & $8.6{\scriptstyle\,\pm\,0.5}$
            & $85.7{\scriptstyle\,\pm\,4.1}$
            & $36.1{\scriptstyle\,\pm\,2.6}$
            & $43.5$ \\
        \midrule
        \multirow{8}{*}{\shortstack{\textsc{Llama}\\ \textsc{3.2-1B}}}
          & Explicit \CoTAcr                 & $58.4$ & $13.9$ & $96.7$ & $65.7$ & $58.8$ \\
          & No-CoT                  & $28.8$ & $ 6.3$ & $50.3$ & $26.7$ & $27.8$ \\
        \cmidrule(l){2-7}
          & CODI      & $52.7$ & $11.9$ & $95.0$ & $60.6$ & $55.8$ \\
          & CODI + SIM-CoT
                                   & $56.1$ & $\mathbf{12.7}$ & $96.2$ & $\mathbf{61.5}$ & $56.8$ \\
          & \textbf{\modelName}
            & $\underline{57.3{\scriptstyle\,\pm\,0.2}}$
            & $\mathbf{12.7{\scriptstyle\,\pm\,0.5}}$
            & $\underline{98.3{\scriptstyle\,\pm\,2.0}}$
            & $60.9{\scriptstyle\,\pm\,0.6}$
            & $\underline{57.3}$ \\
          & \textbf{\modelName + CODI}
            & $\mathbf{58.3{\scriptstyle\,\pm\,0.8}}$
            & $\mathbf{12.7{\scriptstyle\,\pm\,0.2}}$
            & $\mathbf{98.9{\scriptstyle\,\pm\,0.6}}$
            & $\underline{61.1{\scriptstyle\,\pm\,0.9}}$
            & $\mathbf{57.6}$ \\
          & \textbf{\modelName-aux}
            & $55.4{\scriptstyle\,\pm\,0.2}$
            & $12.6{\scriptstyle\,\pm\,0.2}$
            & $97.8{\scriptstyle\,\pm\,0.0}$
            & $58.0{\scriptstyle\,\pm\,0.6}$
            & $56.1$ \\
          & \textbf{\modelName-aux + CODI}
            & $50.6{\scriptstyle\,\pm\,4.7}$
            & $11.4{\scriptstyle\,\pm\,0.8}$
            & $95.7{\scriptstyle\,\pm\,1.2}$
            & $55.0{\scriptstyle\,\pm\,3.6}$
            & $54.0$ \\
        \midrule
        \multirow{8}{*}{\shortstack{\textsc{Llama}\\ \textsc{3.2-3B}}}
          & Explicit \CoTAcr                & $71.5$ & $17.0$ & $98.3$ & $71.0$ & $62.1$ \\
          & No-CoT                  & $38.3$ & $ 9.5$ & $88.7$ & $52.9$ & $50.4$ \\
        \cmidrule(l){2-7}
          & CODI      & $60.8$ & $14.3$ & $98.7$ & $73.3$ & $62.1$ \\
          & CODI + SIM-CoT
                                   & $62.3$ & $14.6$ & $98.8$ & $\underline{74.9}$ & $62.8$ \\
          & \textbf{\modelName}
            & $70.0{\scriptstyle\,\pm\,0.9}$
            & $16.0{\scriptstyle\,\pm\,0.5}$
            & $\mathbf{99.9{\scriptstyle\,\pm\,0.3}}$
            & $\mathbf{75.7{\scriptstyle\,\pm\,0.8}}$
            & $\mathbf{63.9{\scriptstyle\,\pm\,0.3}}$ \\
          & \textbf{\modelName + CODI}
            & $\mathbf{70.6{\scriptstyle\,\pm\,0.2}}$
            & $16.3{\scriptstyle\,\pm\,0.3}$
            & $99.6{\scriptstyle\,\pm\,0.3}$
            & $74.4{\scriptstyle\,\pm\,1.4}$
            & $63.4{\scriptstyle\,\pm\,0.3}$ \\
          & \textbf{\modelName-aux}
            & $69.9{\scriptstyle\,\pm\,0.9}$
            & $\mathbf{16.6{\scriptstyle\,\pm\,0.5}}$
            & $\underline{99.8{\scriptstyle\,\pm\,0.3}}$
            & $74.6{\scriptstyle\,\pm\,2.2}$
            & $\underline{63.7{\scriptstyle\,\pm\,0.8}}$ \\
          & \textbf{\modelName-aux + CODI}
            & $\underline{70.5{\scriptstyle\,\pm\,0.7}}$
            & $\underline{16.5{\scriptstyle\,\pm\,0.4}}$
            & $99.4{\scriptstyle\,\pm\,0.6}$
            & $73.9{\scriptstyle\,\pm\,0.1}$
            & $63.3{\scriptstyle\,\pm\,0.4}$ \\
        \bottomrule
    \end{tabular}
\end{table}

\paragraph{\modelName{} reasons $2.5\times$ faster than explicit \CoTAcr.}
The thought phase dominates the difference between methods (\cref{tab:timing}): \CoTAcr decodes its chain autoregressively, while \modelName{} compresses the same ``thinking'' into $\numLoopPasses$ parallel latent iterations, making its thought $\mathbf{2.5\times}$ faster than \CoTAcr and $\mathbf{1.2\times}$ faster than SIM-CoT\footnote{We measure latency on a single NVIDIA H100 NVL at batch size $1$ with greedy decoding, split into query-prefill, thought, and answer phases. SIM-CoT and CODI both replace each explicit \CoTAcr step with a single latent token decoded sequentially. SIM-CoT additionally adds LoRA adapters (rank $128$) on top of CODI, which inflates every phase.} ($\mathbf{2.1\times}$ faster than \CoTAcr in total\footnote{\modelName{}'s prefill and answer phases are just the standard LM operations (\cref{fig:dslt}), so they match explicit \CoTAcr in \cref{tab:timing} ($16.7$ vs.\ $15.9$~ms and $31.5$ vs.\ $29.5$~ms); the looped latent thought is the only phase that differs.}).
CODI is faster than \modelName{} in the thought phase because it decodes only a single latent per step, rather than \modelName{}'s $\blockBudget\tokPerBlock\!=\!150$ parallel positions, but comes at the cost of accuracy. \modelName{} is nonetheless only modestly slower ($133$ vs.\ $88$~ms) because it consumes all positions in parallel. The inference-time width sweep in \cref{tab:c-thought-infer} further separates sequential depth from parallel width: with $\numLoopPasses$ fixed, increasing the latent prefix from $6$ to $300$ positions changes thought latency by only $30$~ms. 

\begin{table}[t]
    \centering
    \small
    \caption{Per-phase inference latency (average ms/example) on the GSM8K test set with \textsc{Llama-3.2-3B-Instruct} backbone. }
    \label{tab:timing}
    \begin{tabular}{l c c c c}
        \toprule
        Phase & \textbf{\modelName} & Explicit CoT & SIM-CoT & CODI \\
        \midrule
        Query prefill        & $16.7$  & $15.9$   & $29.5$  & $16.5$ \\
        Thought              & $133.0$ & $338.8$  & $162.7$ & $88.2$ \\
        Answer               & $31.5$  & $29.5$   & $59.2$  & $30.4$ \\
        \midrule
        {Sum}         & {$181.2$} & {$384.2$} & {$251.4$} & {$135.1$} \\
        \bottomrule
    \end{tabular}
\end{table}

\paragraph{Speedup grows to $6.9 \times$ on natural-language \CoTAcr traces.}
The main GSM8K-Aug setting uses compact math-expression steps, while natural-language \CoTAcr requires explicit models to decode much longer reasoning in natural language. In the 3B natural-language \CoTAcr stress test (\cref{tab:nl-cot-main}), \modelName{} is on par with explicit \CoTAcr in accuracy ($68.13$ vs.\ $68.41$) while cutting thought-phase latency from $963.6$ ms to $140.8$ ms with a $6.9\times$ speedup. On this natural-language setting, \modelName{} ($68.13\%$ GSM8K) far exceeds the latent baselines \PCCoTAcr ($47.6\%$), \CODIAcr ($55.9\%$), and \KaVaAcr ($60.0\%$) reported by \citet{kuzina2026kava} (\cref{tab:nl-cot-kava-baselines}), staying within variance of explicit \CoTAcr (\cref{tab:nl-cot-main}).

\begin{table}[t]
    \centering
    \small
    \caption{Natural-language \CoTAcr stress test on GSM8K with the \textsc{Llama-3.2-3B-Instruct}. \modelName{} is statistically on par with explicit \CoTAcr accuracy while reducing thought-phase latency by $6.9\times$.}
    \label{tab:nl-cot-main}
    \begin{tabular}{l c c}
        \toprule
        Method & GSM8K acc.\ (\%) $\uparrow$  & Thought ms/example $\downarrow$  \\
        \midrule
        Explicit \CoTAcr & $68.41 \pm 0.59$ &  $963.6$ ($1.0\times$) \\
        \textbf{\modelName} & $68.13 \pm 0.77$ & $\mathbf{140.8}$ ($\mathbf{6.9\times}$)  \\
        \bottomrule
    \end{tabular}
\end{table}

%% file: ablation.tex

\subsection{Ablation Studies}
\label{sec:exp:ablation}

We ablate each component of \modelName to isolate its contribution. Unless noted,
all ablations use \textsc{Llama-3.2-3B-Instruct}, evaluate on
GSM8K test set~\citep{cobbe2021trainingverifierssolvemath}.

\subsubsection{Latent Supervision Design}
\label{sec:exp:ablation:supervision}

\modelName supervises the latent blocks via a direct readout ($\stepLoss$) of the \emph{post-loop} latents $\loopHidden{\numLoopPasses}$ through the main LM head $\lmhead$. We compare it against five alternatives, holding the rest fixed ($\blockBudget{=}\numLoopPasses{=}6$, $\tokPerBlock{=}25$, also keeping $\ansLoss$): \textbf{(i)} no latent supervision (only $\ansLoss$); \textbf{(ii)} CODI~\citep{shen-etal-2025-codi}, distilling a single pre-answer latent onto the teacher CoT model's hidden state; \textbf{(iii)} \modelName{} (per-iter), reading $\loopHidden{t}$ through $\lmhead$ at each iteration $t$ rather than the post-loop $\loopHidden{\numLoopPasses}$; \textbf{(iv)} \modelName-aux, reading $\loopHidden{t}$ through a separate auxiliary decoder $\auxdec$ per iteration (default, \cref{sec:method:lotus-aux}); and \textbf{(v)} \modelName-aux (post-loop), reading $\loopHidden{\numLoopPasses}$ through $\auxdec$ instead.

\paragraph{Both the looped backbone and latent supervision are beneficial.}
Without any latent supervision, the looped backbone alone ($63.3\%$) already exceeds CODI~+~SIM-CoT ($62.3\%$, \cref{tab:accuracy}), which uses a single latent per step; the looped padded backbone is thus already beneficial on its own.
Every supervised variant exceeds both no latent supervision and CODI-only, confirming that the gold \CoTAcr supervision contributes to the gain.
\modelName{} (post-loop direct, $70.0\%$) matches the best \modelName-aux{} schedule (per-iter, $69.9\%$).
The direct (\modelName) and auxiliary decoder (\modelName-aux) routings thus perform comparably at their best schedule.

\paragraph{The best supervision schedule differs between \modelName and \modelName{}-aux.}
\modelName{} performs best with post-loop readout ($70.0\%$ vs.\ $68.2\%$ per-iter), which lets the early latent blocks refine fully until the end of the loop.
\modelName{}-aux instead works best with per-iteration readout ($69.9\%$ vs.\ $68.4\%$ post-loop), which shortens the gradient path---possibly helpful because its extra decoder already lengthens it.


\begin{table}[t]
    \centering
    \small
    \caption{Latent supervision design comparison.}
    \label{tab:supervision-ablation}
    \begin{tabular}{l l c}
        \toprule
        Latent supervision & Routed via & Test acc.\ (\%) $\uparrow$ \\
        \midrule
        None (only $\ansLoss$)              & ---                & $63.3$ \\
        CODI only & --- & $64.4$\\
        \textbf{\modelName} (per-iter supervision of $\loopHidden{t}$)     & main LM head       & $68.2$ \\
        \textbf{\modelName} (post-loop supervision of $\loopHidden{\numLoopPasses}$, default) & main LM head & $\mathbf{70.0}$ \\
        \textbf{\modelName-aux} (per-iter supervision of  $\loopHidden{t}$, default)       & auxiliary decoder  & $69.9$ \\
        \textbf{\modelName-aux} (post-loop supervision of $\loopHidden{\numLoopPasses}$)               & auxiliary decoder  & $68.4$ \\
        \bottomrule
    \end{tabular}
\end{table}

\subsubsection{Looped Architecture Design}
\label{sec:exp:ablation:loop}
\begin{wraptable}[21]{r}{0.32\linewidth}
    \centering
    \footnotesize
    \vspace{-\baselineskip}
    \caption{Training-time looped iterations $\numLoopPasses$ (separate models, latent prefix fixed at $Kc{=}150$).}
    \label{tab:L-ablation}
    \begin{tabular}{@{}cc@{}}
        \toprule
        $\numLoopPasses$ & GSM8K acc.\ (\%) $\uparrow$ \\
        \midrule
        $2$ & $14.6$ \\
        $3$ & $23.2$ \\
        $4$ & $52.6$ \\
        $5$ & $68.1$ \\
        $6$ & $70.0$ \\
        \bottomrule
    \end{tabular}

    \vspace{0.6em}
    \caption{Inference-time looped iterations $\numLoopPasses$ (reusing the model trained at $\numLoopPasses{=}6$).}
    \label{tab:L-infer}
    \begin{tabular}{@{}cc@{}}
        \toprule
        $\numLoopPasses$ & GSM8K acc.\ (\%) $\uparrow$ \\
        \midrule
        $1$ & $22.7$ \\
        $2$ & $40.0$ \\
        $3$ & $55.0$ \\
        $4$ & $63.5$ \\
        $5$ & $68.7$ \\
        $6$ & $70.0$ \\
        $7$ & $69.3$ \\
        \bottomrule
    \end{tabular}
    \vspace{-\baselineskip}
\end{wraptable}

\paragraph{Latent token budget per block \texorpdfstring{$c$}{c}.}
We train separate models at $c\in\{1,5,10,25,30\}$ to measure how the per-block budget affects accuracy, holding $K{=}6$ fixed (\cref{tab:c-thought-ablation}).
Accuracy rises sharply from $c{=}1$ ($49.7\%$) to $c{=}5$ ($67.5\%$), then climbs only marginally before plateauing with $c{=}25$ and $c{=}30$ tied at $70.0\%$.
A single token per CoT step is too narrow for the direct-readout supervision, but moderate widths suffice.

\paragraph{Inference-time \texorpdfstring{$c$}{c}.}
We sweep the inference-time budget $c\in\{1,5,10,25,30,50\}$ on a single checkpoint trained at $c{=}25$ (\cref{tab:c-thought-infer}). Reducing $c$ below the trained value
hurts accuracy ($-19$~points at $c{=}1$, $-11$~points at $c{=}5$, $-6$~points at $c{=}10$),
while exceeding it slightly helps before plateauing ($70.5\%$ at both $c{=}30$ and $c{=}50$).
Thought-phase latency increases only ${\sim}30$~ms
($111$ to $141$~ms) as $c$ varies by $50\times$, since the latents are processed in parallel within the fixed $\numLoopPasses{=}6$ sequential steps.
This contrasts with autoregressive CoT, whose latency grows linearly in the number of thought tokens.

\begin{table}[t]
    \centering
    \small
    \caption{Trained variants of \modelName{} with different latent token budgets per block $c$.}
    \label{tab:c-thought-ablation}
    \begin{tabular}{c c c}
        \toprule
        $c$ (tokens / block) & Total latent positions ($Kc$) & Test acc.\ (\%) \\
        \midrule
        $1$ & $6$ & $49.7$\\
        $5$  & $30$  & $67.5$ \\
        $10$ & $60$  & $68.4$ \\
        $25$ & $150$ & $70.0$ \\
        $30$ & $180$ & $70.0$ \\
        \bottomrule
    \end{tabular}
\end{table}

\begin{table}[t]
    \centering
    \small
    \caption{Inference-time latent budget adaptation $c$
    on \modelName{} trained with $c{=}25$.
    }
    \label{tab:c-thought-infer}
    \begin{tabular}{l c c c}
        \toprule
        Inference-time $c$ &  $Kc$ & GSM8K acc.\ (\%) $\uparrow$ & Avg.\ thought (ms / example) $\downarrow$ \\
        \midrule
        $1$  & $6$   & $51.4$ & $110.9$ \\
        $5$  & $30$  & $59.4$ & $120.3$ \\
        $10$ & $60$  & $64.2$ & $127.0$ \\
        $25$ & $150$ & $70.0$ & $133.0$ \\
        $30$ & $180$ & $70.5$ & $136.0$ \\
        $50$ & $300$ & $70.5$ & $141.2$ \\
        \bottomrule
    \end{tabular}
\end{table}

\paragraph{Looped iterations \texorpdfstring{$\numLoopPasses$}{R}.}

We train separate models for $\numLoopPasses\in\{2,\dots,6\}$, fixing $K{=}6$ and $c{=}25$ so that the latent prefix has $Kc=150$ positions and only the depth of sequential refinement changes (\cref{tab:L-ablation}).
Larger $\numLoopPasses$ gives the model more refinement iterations before the readout.
As a result, accuracy rises steeply with $\numLoopPasses$ (from $14.6\%$ at $\numLoopPasses{=}2$ to $70.0\%$ at $\numLoopPasses{=}6$), with gains continuing through $\numLoopPasses{=}5$ ($68.1\%$) before nearly saturating at $\numLoopPasses{=}6$.

\paragraph{Inference-time \texorpdfstring{$\numLoopPasses$.}{R}}
Reusing the model trained at $\numLoopPasses{=}6$, we vary $\numLoopPasses\in\{1,\dots,7\}$ at inference with the latent prefix fixed at $150$ positions (\cref{tab:L-infer}).
Accuracy climbs with $\numLoopPasses$ ($22.7\%$ at $\numLoopPasses{=}1$ to $70.0\%$ at $\numLoopPasses{=}6$), peaks at the trained $\numLoopPasses{=}6$, and dips slightly at $\numLoopPasses{=}7$ ($69.3\%$).
Running fewer iterations leaves the model less opportunity to refine, while running more than at training time brings no further gain.

%% file: latent_analysis.tex
\section{Latent Representation Analysis}
\label{sec:exp:analysis}

We analyze the learned latent representations along three axes. Together, these analyses show that \modelName{}'s accuracy gains are accompanied by structured latent computation: the post-loop latents carry readable \CoTAcr token signal, assign nontrivial mass to unseen valid alternatives, and rely on both gold \CoTAcr supervision and answer supervision to form coherent reasoning states. \Cref{sec:exp:analysis:cross-reader} measures the likelihood assigned to the gold \CoTAcr under different readouts, \cref{sec:exp:analysis:multipath} tests whether the latents place mass on unseen but valid reasoning chains, and \cref{sec:exp:analysis:loss-ablation} ablates the roles of gold \CoTAcr and answer supervision in shaping the representation. Extended block-to-step decode examples are deferred to \cref{app:greedy-examples}.

\subsection{Likelihood of the Gold Chain under Different Readout Options}
\label{sec:exp:analysis:cross-reader}

Recall that $\loopHidden{\numLoopPasses}$ denotes the post-loop latents and $\auxHidden{t}$ the aux decoder's output latents for $t\in\{1,\dots,\numLoopPasses\}$. We evaluate four readouts in~\cref{tab:joint-logp-cross}: \textbf{(i)} \modelName{}, reading $\loopHidden{\numLoopPasses}$ through the base LM head; \textbf{(ii)} \modelName-aux{}, reading $\loopHidden{\numLoopPasses}$ through $\auxhead$; \textbf{(iii)} \modelName-aux{}, reading $\auxHidden{t}$ through $\auxhead$ under \emph{teacher forcing} (TF) that matches training (each block's $\tokPerBlock$ latents paired with that block's gold step tokens); and \textbf{(iv)} \modelName-aux{}, reading $\auxHidden{t}$ through $\auxhead$ but under \emph{free-running} (FR),\footnote{\modelName-aux{} (FR) gives a fairer reference to \modelName{} than \modelName-aux{} (TF): at inference the gold \CoTAcr is unavailable, so \modelName-aux{} must feed its decoder its own generated tokens, exactly the FR setting. The TF row instead feeds the gold prefix the model would not have at test time.} feeding the aux decoder's own generated token in place of the gold tokens.\footnote{Here we report greedy decoding results. Replacing them with temperature-$T$ sampling ($T \in \{0.5, 0.7, 1.0, 1.5\}$) moves FR NLL by $\leq 0.05$ and top-$1$ by $\leq 1$ pp.}

\paragraph{Metrics.}
We report the negative log-likelihood (NLL) of the gold \CoTAcr token sequence $\thoughtSeq$ at the latent positions.\footnote{Note that the $\loopHidden{\numLoopPasses}$ readout scores the gold \CoTAcr under the PCL (\cref{sec:method:theory}) factorization and $\auxHidden{t}$ under an autoregressive one, so NLL is only directly comparable under the same factorization and only for reference across factorizations. Mathematically, PCL NLL has a higher lower bound than the autoregressive NLL (details in \cref{app:pcl-bound}).}
We also report average ${P(\text{gold}\in\text{top-1})}$ and ${P(\text{gold}\in\text{top-5})}$ against the full $128\text{K}$-vocabulary softmax. All metrics are averaged over the $1{,}319$ GSM8K test questions. 

\paragraph{\modelName's post-loop latent $\loopHidden{\numLoopPasses}$ contains the gold \CoTAcr via a direct readout.}
Reading \modelName{}'s post-loop latent $\loopHidden{\numLoopPasses}$ through $\lmhead$ assigns the gold \CoTAcr a low $3.07$ NLL and $70.9\%$ top-$1$ accuracy (\cref{tab:joint-logp-cross}), edging out \modelName-aux{} (FR) at $64.5\%$. 
The teacher-forced \modelName-aux{} reader does better still ($1.56$ NLL, $84.3\%$ top-$1$), but with the caveat that it uses the gold prefix as input, so we read it as an upper bound rather than a direct comparison. Still, the results confirm that the latents carry enough information to reconstruct the gold \CoTAcr chain.

\paragraph{\modelName{}-aux has \CoTAcr signal in both $\loopHidden{\numLoopPasses}$ and $\auxHidden{t}$.}
Surprisingly, even though $\loopHidden{\numLoopPasses}$ is never directly supervised to predict the gold \CoTAcr through $\auxhead$ in \modelName-aux{}, we can still read $\loopHidden{\numLoopPasses}$ through $\auxhead$ with a $25.8\%$ top-$5$ probability.
Greedy decoding also surfaces some meaningful gold tokens, indicating that the \CoTAcr content is carried by the latents themselves. In fact, despite being supervised through different heads, \modelName{} and \modelName-aux{} have a \emph{similar} effect of making the post-loop latents $\loopHidden{\numLoopPasses}$ put mass on the \CoTAcr steps, with the caveat that \modelName-aux{} readouts are less formatted (which is expected since it is not supervised directly). We provide side-by-side $\loopHidden{\numLoopPasses}$ readout examples for both models in \cref{app:latent-examples}.

\begin{table}[t]
    \centering
    \small
    \caption{Likelihood of the gold \CoTAcr from different readout configurations.
    }
    \label{tab:joint-logp-cross}
    \begin{tabular}{l r r r}
        \toprule
        Reader & NLL  & $P(\text{gold}\in\text{top-1})$ & $P(\text{gold}\in\text{top-5})$ \\
        \midrule
        \multicolumn{4}{l}{\emph{Without gold prefix:}} \\
        \modelName{}: $\loopHidden{\numLoopPasses}$ through $\lmhead$                       & $3.07$  & $70.9\%$ & $85.8\%$ \\
        \modelName-aux: $\loopHidden{\numLoopPasses}$ through $\auxhead$              & $9.02$  & $12.4\%$ & $25.8\%$ \\
        \modelName-aux: $\auxHidden{t}$ through $\auxhead$ (FR)                       & $6.39$  & $64.5\%$ & $76.7\%$ \\
        \midrule
        \multicolumn{4}{l}{\emph{With gold prefix:}} \\
        \modelName-aux: $\auxHidden{t}$ through $\auxhead$ (TF)                       & $1.56$  & $84.3\%$ & $93.9\%$ \\
        \bottomrule
    \end{tabular}
\end{table}

\subsection{Mass on Unseen Valid Chains}
\label{sec:exp:analysis:multipath}

Beyond the gold chain in the test set, we test whether \modelName{} and \modelName-aux{} place mass on unseen valid reasoning chains rather than only memorizing the seen trace in the training set. We use the same four readout configurations as in \cref{sec:exp:analysis:cross-reader}.

\paragraph{Setup.}
Given each question, we compare its ground-truth reasoning chain against an unseen-but-valid alternative. Let $G$ and $U$ denote the sets of intermediate numbers appearing uniquely in the ground-truth and the unseen reasoning chain, respectively.
A random control set serves as a baseline. We report each set's average per-token negative log-likelihood (NLL) at the position where it surfaces most strongly (lower means more confident). We also report $P(U\in\text{top-}k)$, the fraction of $U$ numbers that surface within the top-$k$ readout at any latent position. See \cref{app:multipath-setup} for the full construction and metric details.

\paragraph{Findings.}
Across all four readouts the ordering $G \!\ll\! U \!\ll\!$ Random holds (\cref{tab:path-mass}, lower NLL $=$ surfaces more): each reader assigns much lower NLL to a ground-truth-only number than to an unseen-but-valid one, which in turn sits far below the random control. Crucially, the $U$ signal is concentrated rather than diffuse: $P(U\in\text{top-}k)$ columns confirm the unseen-valid intermediates genuinely surface in the top-$k$ (e.g., for \modelName{} through the base LM head, {$15.3\%$} are top-$1$ and {$64.0\%$} fall within top-$5$, despite never being trained on). \Cref{app:multipath-example} and \cref{fig:multipath-grid6} give a path-level example and visual grid: two valid chains reach the same answer ($108$) through disjoint intermediates, and the unseen-path numbers ($24$, $18$) surface in the post-loop readout despite never appearing in the trained chain, question, or answer.

\subsection{Disentangling the Two Losses in \modelName}
\label{sec:exp:analysis:loss-ablation}

We now ask what each of \modelName's two supervision losses contributes to the latent representation observed above. The theory in \cref{sec:method:theory} ascribes complementary roles to $\stepLoss$ and $\ansLoss$. We compare three models that differ in supervision: \modelName{} (both losses), $\stepLoss$ only, and $\ansLoss$ only.

\paragraph{Gold-chain NLL.}
Reading the $\loopHidden{\numLoopPasses}$ through the base LM head $\lmhead$, \Cref{tab:joint-logp} reports NLL, $P(\text{gold}\in \text{top-1})$, and $P(\text{gold}\in \text{top-5})$ on the GSM8K test set, using the same protocol as \cref{tab:joint-logp-cross}.

\begin{table}[t]
    \centering
    \small
    \caption{Evaluation on both seen and unseen valid \CoTAcr chains.}
    \label{tab:path-mass}
    \begin{tabular}{l c c c c c}
        \toprule
        Reader & $G$ NLL & $U$ NLL & Random NLL & $P(U\in \text{top-1})$ & $P(U\in \text{top-5})$ \\
        \midrule
        \multicolumn{6}{l}{\emph{Without gold prefix:}} \\
        \modelName{}: $\loopHidden{\numLoopPasses}$ through $\lmhead$              & $0.07$ & {$4.28$} & {$8.16$} & {$15.3\%$} & {$64.0\%$} \\
        \modelName-aux: $\loopHidden{\numLoopPasses}$ through $\auxhead$            & {$3.30$} & {$6.04$} & {$9.42$} & {$14.5\%$} & {$49.3\%$} \\
        \modelName-aux: $\auxHidden{t}$ through $\auxhead$ (FR)                     & {$0.80$} & {$12.62$} & {$20.39$} & {$5.6\%$}  & {$39.2\%$} \\
        \midrule
        \multicolumn{6}{l}{\emph{With gold prefix:}} \\
        \modelName-aux: $\auxHidden{t}$ through $\auxhead$ (TF)                     & $0.18$ & {$13.17$} & {$20.83$} & {$1.9\%$}  & {$37.4\%$} \\
        \bottomrule
    \end{tabular}
\end{table}

\begin{table}[t]
    \centering
    \small
    \caption{Disentangling on the two loss components; NLL and per-position gold \CoTAcr accuracy on the GSM8K test.}
    \label{tab:joint-logp}
    \begin{tabular}{l c c c}
        \toprule
        Variant & NLL $\downarrow$ & $P(\text{gold}\in \text{top-1})$ $\uparrow$ & $P(\text{gold}\in \text{top-5})$ $\uparrow$ \\
        \midrule
        \modelName                                   & $3.07$ & $70.9\%$ & $85.8\%$ \\
        $\stepLoss$ only            & $9.29$ & $9.1\%$  & $17.1\%$ \\
        $\ansLoss$ only             & $5.97$ & $9.4\%$  & $26.5\%$ \\
        \bottomrule
    \end{tabular}
\end{table}

\paragraph{$\stepLoss$ and $\ansLoss$ play complementary roles.}
\modelName{} beats both ablations (\cref{tab:joint-logp}). $\stepLoss$-only, lacking the cross-block coupling and the answer grounding that  $\ansLoss$ supplies, is worst on every column. $\ansLoss$-only, without directly tying $\loopHidden{\numLoopPasses}$ to the gold \CoTAcr, lands nearer the gold tokens (NLL $5.97$, top-$5$ $26.5\%$) yet has a top-$1$ rate similar to $\stepLoss$-only and stays far short of \modelName{} (NLL $3.07$, top-$5$ $85.8\%$). The answer loss alone settles $\loopHidden{\numLoopPasses}$ into an answer-consistent neighborhood without aligning to the gold chain, matching the complementary roles posited in \cref{sec:method:theory}.

%% file: conclusion.tex
\section{Conclusion}
\label{sec:conclusion}
We introduce \modelName, showing that latent reasoning can approach the performance of explicit \CoTAcr by supervising a looped padded Transformer in parallel against the gold \CoTAcr tokens under the simple cross-entropy objective. On \textsc{Llama-3.2-3B-Instruct}, \modelName bridges the in-domain gap to explicit \CoTAcr on GSM8K, surpasses \CoTAcr on the out-of-domain average, and cuts thought-phase latency by $2.5\times$. Ablations show the looped backbone, parallel gold \CoTAcr supervision, and sufficient block width and loop depth are each necessary. The latent representation analysis further shows the latents are transparent: the gold \CoTAcr is recoverable from them by a direct readout, they place graded probability on unseen but valid reasoning chains rather than a single memorized trace, and the step and answer losses contribute complementary structure.

\paragraph{Limitations.}
We follow prior latent reasoning work~\citep{hao2025training,shen-etal-2025-codi,wei2025simcotsupervisedimplicitchainofthought} and evaluate on math benchmarks. Whether the recipe transfers to other domains remains an open direction for future work.
The block budget $\blockBudget$, per-block width $\tokPerBlock$, and loop depth $\numLoopPasses$ are fixed hyperparameters that must be set to cover the expected step count, so chains longer than $\blockBudget$ steps fall back to autoregressive completion of the tail.
Making $\blockBudget$, $\tokPerBlock$, and $\numLoopPasses$ adaptive is a natural and promising direction for future work.

%% file: appendix.tex

\appendix
\crefalias{section}{appendix}
\crefalias{subsection}{appendix}
\crefalias{subsubsection}{appendix}
\newpage
\clearpage
{\LARGE \textbf{Appendix}} \par
\startcontents[sections]
\vspace{0.5em}
\noindent\Cref{app:related} discusses related works including looped transformers, latent reasoning, and speculative decoding methods. \Cref{app:notation} summarizes the notation used in the paper. \Cref{app:differentiation} compares \modelName{} and \modelName-aux with the closest latent reasoning baselines. \Cref{app:details} includes the experimental and implementation details. \Cref{app:pcl-bound} explains the relation between PCL and autoregressive NLL lower bounds. Finally, \Cref{app:latent-examples} provides the latent representation analysis details, including greedy readout examples and the multi-path setup and examples.\par
\vspace{0.5em}
\printcontents[sections]{ }{1}{}
\clearpage
\input{related_appendix}

\section{Notation}
\label{app:notation}
\Cref{tab:notation} summarizes the notation used throughout the paper.
\input{notation.tex}
\section{Closely Related Latent Reasoning Methods}
\label{app:differentiation}

We position \modelName and its auxiliary-decoder variant \modelName-aux against the most closely related latent reasoning methods.

\paragraph{Coconut~\citep{hao2025training}} introduced a curriculum-based approach that progressively replaces explicit \CoTAcr steps with latent tokens conditioned autoregressively on the previous latent. Both \modelName and \modelName-aux depart from Coconut in two ways: 
\begin{enumerate*}[label=\textit{(\roman*)}]
    \item they use a \emph{padded} latent prefix processed by a \emph{looped} Transformer, so all $\blockBudget$ steps are decoded in parallel, not autoregressively, giving the $\bigOFun{\numLoopPasses}$-iteration thought-phase cost, and
    \item they add step-aligned supervision on the latent blocks, where Coconut supervises only the answer.
\end{enumerate*}
Our ablation without latent supervision (\cref{tab:supervision-ablation}) is the Coconut-style regime within our looped architecture, and it underperforms both variants.

\paragraph{CODI~\citep{shen-etal-2025-codi} and SIM-CoT~\citep{wei2025simcotsupervisedimplicitchainofthought}} retain Coconut's autoregressive latent decoding but add an auxiliary supervision target: CODI aligns the hidden state of a single designated token with the corresponding teacher hidden state from a CoT model, and SIM-CoT trains an auxiliary autoregressive decoder that aligns each latent token with the corresponding \CoTAcr token. Our \modelName removes the auxiliary path entirely, supervising the model's own post-loop hidden states through the answer LM head at block granularity. \modelName-aux keeps an auxiliary autoregressive decoder, but produces all blocks in parallel through the looped backbone and supervises a full block per step rather than one latent per step.

\paragraph{Parallel Continuous CoT (PCCoT)~\citep{wu-etal-2025-parallel} and KaVa~\citep{kuzina2026kava}}
form a single Jacobi-iteration family: \PCCoTAcr introduces the backbone and \KaVaAcr builds on it. They both refine a fixed budget of $24$ latent tokens over $3$ Jacobi iterations and differ only in supervision. Like \modelName{}, they process latent symbols in parallel and refine them over passes, but differ from our methods in three ways:
\begin{enumerate*}[label=\textit{(\roman*)}]
  \item \emph{Architecture}: \PCCoTAcr and \KaVaAcr use right-shifted Jacobi iterations to simulate sequential continuous \CoTAcr in parallel~\citep{hao2025training}, so their latents are continuous thought vectors with no fixed alignment to individual gold \CoTAcr steps. \modelName{} instead pins block $\blockidx$ to fixed positions across all loops, so it always maps to gold \CoTAcr step $\blockidx$ and is naturally supervised against that step's tokens;
  \item \emph{Supervision}: \PCCoTAcr uses CODI~\citep{shen-etal-2025-codi}, and \KaVaAcr supervises indirectly by distilling a compressed teacher KV-cache, whereas both our variants supervise each latent block against its exact gold \CoTAcr step; and
  \item \emph{Depth}: \PCCoTAcr reports accuracy peaking at about $3$ Jacobi iterations and degrading beyond that, which is why \KaVaAcr inherits the $3$-iteration setting, whereas \modelName{} improves monotonically with loop depth up to 6 (\cref{tab:L-ablation}).
\end{enumerate*}

\begin{table}[t]
    \centering
    \small
    \caption{Comparison with latent reasoning methods in other step regimes on GSM8K (trained on GSM8k-Aug). }
    \label{tab:pccot-kava}
    \begin{tabular}{l c c c}
        \toprule
        Method & \textsc{GPT-2} & \textsc{Llama-3.2-1B} & \textsc{Llama-3.2-3B} \\
        \midrule
        Explicit \CoTAcr & $42.7$ & $58.4$ & $71.5$ \\
        \midrule
        \ccntAcr~\citep{hao2025training}        & $36.6$ & $33.2$ & $-$ \\
        \ccntAcr + SIM-CoT~\citep{wei2025simcotsupervisedimplicitchainofthought} & $44.8$ & $42.2$ & $-$ \\
        \midrule
        \PCCoTAcr~\citep{wu-etal-2025-parallel} & $\mathbf{49.5}$ & $54.2$ & $54.7$ \\
        \KaVaAcr~\citep{kuzina2026kava}         & $-$ & $56.5$ & $65.7$ \\
        \textbf{\modelName}                     & $44.1$ & $\mathbf{57.3}$ & $\mathbf{70.0}$ \\
        \bottomrule
    \end{tabular}
\end{table}

\paragraph{Comparison with latent methods at other compute budgets.} The main results (\cref{tab:accuracy}, \cref{sec:exp:main}) compare \modelName{} against latent methods at the \emph{same} sequential step budget (CODI and CODI + SIM-CoT, both $\numLoopPasses{=}6$). Here we provide a more comprehensive accuracy comparison in \cref{tab:pccot-kava} against latent methods whose sequential compute budget instead differs from \modelName's $\numLoopPasses{=}6$ sequential steps. \ccntAcr and \ccntAcr + SIM-CoT decode latents autoregressively under the curriculum of \citet{hao2025training} ($10$ autoregressive latent tokens), while \PCCoTAcr and \KaVaAcr decode in parallel over a shared Jacobi-iteration backbone ($24$ latent tokens in total, refined over $3$ iterations). 
We take accuracies from \citet{wei2025simcotsupervisedimplicitchainofthought}, \citet{kuzina2026kava}, and \citet{wu-etal-2025-parallel} for each method. \modelName{} attains higher in-domain GSM8K accuracy than \KaVaAcr at both shared scales ($57.3$ vs.\ $56.5$ at $1$B, $70.0$ vs.\ $65.7$ at $3$B), staying within $1.5$ points of explicit \CoTAcr while \PCCoTAcr and \KaVaAcr fall further behind as the backbone grows. The missing entries follow what each source reports: \citet{wei2025simcotsupervisedimplicitchainofthought} evaluate \ccntAcr only up to \textsc{Llama-3.2-1B}, attributing \ccntAcr's failure at larger scale to a latent-instability issue in which the latent representations become homogeneous and lose semantic diversity, causing training to collapse as the latent budget grows. \citet{kuzina2026kava} do not run \KaVaAcr on \textsc{GPT-2}.

\begin{table}[t]
    \centering
    \small
    \caption{{Comparison with latent reasoning methods on GSM8K, trained on
GSM8k-Aug (the natural language version in \citet{wu-etal-2025-parallel}).} }
    \label{tab:nl-cot-kava-baselines}
    \begin{tabular}{l c c c}
        \toprule
        Method & GSM8K acc.\ (\%) $\uparrow$ & GSM-Hard acc.\ (\%) $\uparrow$ & SVAMP acc.\ (\%) $\uparrow$ \\
        \midrule
        Explicit \CoTAcr & $68.41 \pm 0.59$ & $18.27 \pm 0.85$ & $71.93 \pm 1.62$ \\
        \midrule
        \PCCoTAcr & $47.6$ & $11.0$ & $65.2$ \\
        \CODIAcr & $55.9$ & $13.6$ & $70.1$ \\
        \KaVaAcr & $60.0$ & $14.8$ & $66.1$ \\
            \textbf{\modelName}& $\mathbf{68.13 \pm 0.77}$ & $\mathbf{16.27 \pm 0.19}$& $\mathbf{73.40 \pm 0.35}$ \\
        \bottomrule
    \end{tabular}
\end{table}
\paragraph{{Natural-language \CoTAcr stress-test.}}
{For the natural-language \CoTAcr stress test in \cref{sec:exp:main}, we use the \textsc{Llama-3.2-3B-Instruct} backbone. The explicit \CoTAcr run reaches $68.41\%$ GSM8K accuracy with $963.6$ ms thought latency. The looped run reaches $68.13\%$ accuracy with $140.8$ ms thought latency. The latent baselines trail well behind on GSM8K (\PCCoTAcr $47.6$, \CODIAcr $55.9$, \KaVaAcr $60.0$). Out of domain, \modelName{} leads all methods on SVAMP ($73.40$, above explicit \CoTAcr's $71.93$) and stays close to explicit \CoTAcr on GSM-Hard ($16.27$ vs.\ $18.27$), ahead of every latent baseline (\cref{tab:nl-cot-kava-baselines}). } {We exclude SIM-CoT from this comparison because \citet{suleymanzade2026mux} report weak SIM-CoT accuracy on the natural-language subset.}

\section{Experimental Details}
\label{app:details}
This appendix includes the training and evaluation details in \cref{sec:exp:setup}.

\paragraph{Latent prefix budget.} Each of the $\blockBudget$ blocks holds $\tokPerBlock$ latent positions, so the prefix spans $\blockBudget\tokPerBlock$ positions: $\tokPerBlock{=}25$ ($150$ positions) for the \textsc{Llama} backbones and $\tokPerBlock{=}13$ ($78$ positions) for \textsc{GPT-2}. We use $\blockBudget{=}6$, which covers the reference \CoTAcr step count for $99\%$ of GSM8k-Aug examples. Longer problems fall back to autoregressive completion of the unsupervised tail.

\paragraph{Training curriculum.} We follow the staged curriculum of~\citet{hao2025training}, converting one additional \CoTAcr step into a latent block every $E_\text{stage}$ epochs. At epoch $e$, we use
\[
\blockBudget_e \;=\; \min\!\left(\lfloor e / E_\text{stage}\rfloor,\; \blockBudget\right),
\]
latent blocks: up to the first $\blockBudget_e$ \CoTAcr steps become latent blocks, and the rest remain visible tokens. Thus $e{=}0$ is standard \CoTAcr fine-tuning and the curriculum saturates at $\blockBudget_e{=}\blockBudget$. We set the loop count at epoch $e$ to $\numLoopPasses_e = r\,\blockBudget_e$, where $r$ is the loops-per-block ratio. Since each pass refines all blocks in parallel, $\blockBudget_e$ controls the latent workspace width and $\numLoopPasses_e$ controls the depth. Unless stated otherwise, $E_\text{stage}{=}1$ and $r{=}1$, so the saturated model uses $\numLoopPasses{=}\blockBudget{=}6$. The loop-depth ablation (\cref{sec:exp:ablation:loop}) fixes $\blockBudget{=}6$ and varies $r$ to decouple depth from width.

\paragraph{Optimization details.} We initialize from each backbone's standard explicit \CoTAcr checkpoint, and fine-tune all parameters with AdamW and gradient-norm clipping at $1.0$, in \texttt{bf16} for the Llama backbones and \texttt{fp32} for \textsc{GPT-2}. The learning rate is constant: $1\!\times\!10^{-4}$ for \textsc{GPT-2} and \textsc{Llama-3.2-1B-Instruct}, and $5\!\times\!10^{-5}$ for \textsc{Llama-3.2-3B-Instruct}. The training objective (\cref{eq:loss}) weights the step loss at $\stepLossWeight{=}0.05$ (for both \modelName{} and \modelName{}-aux). Adding CODI weights its loss at $1.0$. We train for $30$ epochs with an overall batch size of $128$, and select the best checkpoint by GSM8k validation accuracy, though the final epoch checkpoints perform similarly.
\paragraph{Natural-language training setup.}
For the natural-language stress-test on \modelName, we keep the same 3B looped backbone but switch to the natural-language GSM8k-Aug training set~\citep{wu-etal-2025-parallel}, increase latent width to $c{=}50$ and set $\stepLossWeight$ to $0.033$. We initialize from the CoT checkpoint trained on the natural-language GSM8k-Aug. Other settings are identical to those used for the original GSM8k-Aug training.

\paragraph{\modelName-aux auxiliary decoder.}
For \modelName-aux, the auxiliary decoder is a full-size deep copy of the base model, initialized from its weights but trained together with the main model. This full-size choice matches \SiMCoTAcr~\citep{wei2025simcotsupervisedimplicitchainofthought}, whose auxiliary decoder is architecturally identical to the base model. We also tried lightweight auxiliary decoders but found them unstable in training.

\paragraph{Evaluation.}
Inference uses greedy decoding with batch size $1$. This isolates per-example latency from batching effects and calculates the average latency per sample accurately. Latency is measured on a single NVIDIA H100 and decomposed into query, thought, and answer phases (\cref{tab:timing}).

\paragraph{Batched KV-cache sharing.}
For training, we left-pad shorter examples so the $\latTokenSym$ positions align across the batch. This lets us compute each example's question-prefix KV cache once and reuse it across all $\numLoopPasses$ looped iterations.

\section{PCL Versus Autoregressive NLL Lower Bounds}
\label{app:pcl-bound}

Here we provide the explanation mentioned in \cref{sec:exp:analysis:cross-reader}. PCL (\cref{sec:method:theory}) scores each gold token in parallel, $-\sum_{i,j}\log \lm(\stepToken{\blockidx}{j}\mid\loopHidden{\numLoopPasses})$, whereas the autoregressive factorization also conditions on the preceding gold tokens, $-\sum_{i,j}\log \lm(\stepToken{\blockidx}{j}\mid\loopHidden{\numLoopPasses},\thoughtSeq_{<(\blockidx,j)})$, where $\thoughtSeq_{<(\blockidx,j)}$ denotes all gold tokens before position $(\blockidx,j)$ in reading order. The lowest expected NLL each can reach is the corresponding conditional entropy, and since conditioning never increases entropy,
\begin{equation}
H\!\big(\stepToken{\blockidx}{j} \mid \loopHidden{\numLoopPasses}\big)
\;\geq\;
H\!\big(\stepToken{\blockidx}{j} \mid \loopHidden{\numLoopPasses}, \thoughtSeq_{<(\blockidx,j)}\big),
\label{eq:pcl-ar-ineq}
\end{equation}
with equality only when the gold tokens are conditionally independent given the latents. So PCL has a strictly higher NLL floor whenever cross-token dependence exists.

\section{Details for Latent Representation Analysis\texorpdfstring{ (\cref{sec:exp:analysis})}{}}
\label{app:latent-examples}
\subsection{Greedy Readout Examples}
\label{app:greedy-examples}
Here we give examples of two correctly answered and two incorrectly answered GSM8K test problems. We show three readouts of the post-loop latents: \textbf{(A)} the per-block top-$1$ token sequence reading $\loopHidden{\numLoopPasses}$ through \modelName{}'s base LM head $\lmhead$; \textbf{(B)} the same readout applied to \modelName-aux's post-loop latents $\loopHidden{\numLoopPasses}$ through its auxiliary head $\auxhead$; and \textbf{(C)} the autoregressive \CoTAcr produced by \modelName-aux's trained auxiliary decoder from $\auxHidden{t}$, conditioned on the same latents and decoded greedily from the start-of-CoT token. (A) and (B) show the top-$1$ token stream per block verbatim (newlines rendered as \texttt{\textbackslash n}), truncated at $60$ characters per block with \texttt{...} marking the cut. (C) shows the raw aux-decoder output. We additionally report the final answer the model generates autoregressively from $\loopHidden{\numLoopPasses}$ after all $\numLoopPasses{=}6$ loops. In each \textbf{(A)}/\textbf{(B)} readout we list the $\blockBudget{=}6$ latent blocks top to bottom, labeled \texttt{B0}--\texttt{B5}.

For readability, \rdhl{blue} marks the readable gold content: well-formed \texttt{<<\dots>>} \CoTAcr steps and exact gold chain numbers (step operands and intermediate results), including a number stuttered back-to-back such as \texttt{540540} for $540$. Wrong digit runs (e.g.\ \texttt{120120} or \texttt{3636}) are left unmarked, and within each readout line we highlight only the first surfacing of each gold value.

\paragraph{Example 1 — Janet's ducks.} \quad

\emph{Question: Janet's ducks lay 16 eggs per day. She eats three for breakfast every morning and bakes muffins for her friends every day with four. She sells the remainder at the farmers' market daily for \$2 per fresh duck egg. How much in dollars does she make every day at the farmers' market?} \emph{Reference:} \texttt{<<16-3-4=9>>}, \texttt{<<9*2=18>>}. \emph{Gold answer $18$.} 

\begin{quote}\small
\textbf{(A) \modelName{} LM head} ($\loopHidden{\numLoopPasses}$ through $\lmhead$).
\begin{Verbatim}[commandchars=\@\[\]]
B0: @rdhl[<<3+4=7>>]\n00>>\n7>>\n>>\n>>\n@rdhl[16]16--@rdhl[9]999999<<
B1: @rdhl[<<16-7=9>>]\n00>>\n>>\n>>\n>>\n>>\n*2==@rdhl[18]18>>\n>>\n>>\n1818<<
B2: @rdhl[<<9*2=18>>]\n00>>\n>>\n>>\n18>>\n>>\n>>\n= 1818>>\n18>>\n>>\n...
B3: @rdhl[<<18*2=18>>]\n>>\n>>\n>>\n>>\n>>\n>>\n>>\n>>\n=@rdhl[9]9>>\n>>\n18>>...
B4: ###134*@rdhl[7]=>>\n@rdhl[18]>>\n>>\n>>\n>>\n>>\n>>\n>>\n2=2 .>>\n181818>>...
B5: >>\n@rdhl[18]=21818>>\n>>\n>>\n===.###2== ,>>\n1818222
\end{Verbatim}

\textbf{(B) \modelName-aux LM head} ($\loopHidden{\numLoopPasses}$ through $\auxhead$).
\begin{Verbatim}[commandchars=\@\[\]]
B0: <<'+)@rdhl[7]))<<@rdhl[16]16--7= 1313\n\n###22*22
B1: =2189)#########2222=  @rdhl[18]18#########22*@rdhl[9]=
B2:  @rdhl[18]18 per#########2922 =1818 fresh######@rdhl[9]9* no@rdhl[16]= 18
B3: @rdhl[18] fresh<|eot_id|>#########@rdhl[9]* no no nothing=1818 Clean<|eot_...
B4: @rdhl[18]ly<|eot_id|>###### no nothing no nothing 1818<|eot_id|>###...
B5: ############ nothing nothing nothing=@rdhl[18]1818############### n...
\end{Verbatim}

\textbf{(C) \modelName-aux auxiliary decoder} ($\auxHidden{t}$ through $\auxhead$, FR).\\
\texttt{\rdhl{<<3+4=7>>}\textbackslash n\rdhl{<<16-7=9>>}\textbackslash n\rdhl{<<9*2=18>>}\textbackslash n\rdhl{<<18=18>>}\textbackslash n\rdhl{<<9*2=18>>}\textbackslash n\rdhl{<<18+18=18>>}\textbackslash n\rdhl{<<18*2=18>>}\textbackslash n\rdhl{<<18=18>>}}

\textbf{Final answer (autoregressive, after $6$ loops).} \modelName{}: \texttt{\#\#\# 18}. \modelName-aux: \texttt{\#\#\# 18}. Gold: $18$. Both correct.

\textbf{Reading the chains.} (A): the gold intermediates $\mathbf{9}$ and $\mathbf{18}$ surface from B1 onward and stabilize by B5. (B): the same numerics ($9, 18, 16, 7$) are present but interleaved with \texttt{<|eot\_id|>} runs and filler. (C): the aux decoder produces the gold first two steps \texttt{<<3+4=7>>}, \texttt{<<16-7=9>>}, \texttt{<<9*2=18>>} and then loops on $18$.
\end{quote}

\paragraph{Example 2 — James's sprints.} \quad

\emph{Question: James decides to run 3 sprints 3 times a week. He runs 60 meters each sprint. How many total meters does he run a week? Reference:} \texttt{<<3*3=9>>}, \texttt{<<9*60=540>>}. \emph{Gold answer $540$.}

\begin{quote}\small
\textbf{(A) \modelName{} LM head} ($\loopHidden{\numLoopPasses}$ through $\lmhead$).
\begin{Verbatim}[commandchars=\@\[\]]
B0: @rdhl[<<3*60=180>>]\n00>>\n180180>>\n>>\n3==@rdhl[540]540>>\n>>\n>>\n>>\n>...
B1: @rdhl[<<180*3=540>>]\n00>>\n>>\n>>\n540>>\n###   540>>\n>>\n   >>\n...
B2: @rdhl[<<540*540=540>>]\n00>>\n>>\n>>\n>>\n>>\n>>\n###  540>>\n>>\n ...
B3: @rdhl[<<540=180=540>>]\n540>>\n>>\n>>\n>>\n>>\n>>\n###  540>>\n>>\n...
B4: ###@rdhl[540]= =>>\n###>>\n>>\n>>\n 2 ######  540540>>\n>>\n   ###
B5: ### == >>\n#########    >>\n###    >>\n>>\n   >>\n
\end{Verbatim}

\textbf{(B) \modelName-aux LM head} ($\loopHidden{\numLoopPasses}$ through $\auxhead$).
\begin{Verbatim}[commandchars=\@\[\]]
B0: @rdhl[3]3*3= @rdhl[180]180180*###9993==  @rdhl[540]540)######@rdhl[9]
B1: @rdhl[9]*@rdhl[60]= @rdhl[540]540))#########9*@rdhl[3] nothing = 540540)############
B2: /* nothing= @rdhl[540]540)#########540+### nothing  540############...
B3:  nothing is @rdhl[540]540###############540,### also 540540540#####...
B4:  is@rdhl[540]540540############540,### also 540540#################...
B5: @rdhl[540]540###############540### definitely 540540540############...
\end{Verbatim}

\textbf{(C) \modelName-aux auxiliary decoder} ($\auxHidden{t}$ through $\auxhead$, FR).\\
\texttt{\rdhl{<<3*60=180>>}\textbackslash n\rdhl{<<180*3=540>>}\textbackslash n\rdhl{<<540=540>>}\textbackslash n\rdhl{<<540/60=9>>}\textbackslash n\rdhl{<<540=540>>}\textbackslash n\rdhl{<<540/10=540>>}\textbackslash n\rdhl{<<540>>}}

\textbf{Final answer (autoregressive, after $6$ loops).} \modelName{}: \texttt{\#\#\# 540}. \modelName-aux: \texttt{\#\#\# 540}. Gold: $540$. Both correct.

\textbf{Reading the chains.} (A): the gold step \texttt{<<3*60=180>>} appears in B0 already (a valid alternative decomposition: meters per day first), and $\mathbf{540}$ stabilizes from B1. (B): $180, 540, 9, 3$ are all present but with heavy \texttt{\#\#\#} runs around them. (C): the aux decoder reproduces $\mathbf{180} \to \mathbf{540}$ cleanly, then loops.
\end{quote}

\paragraph{Example 3 — Carla's download (failure).} \quad

\emph{Question: Carla is downloading a 200 GB file. Normally she can download 2 GB/minute, but 40\% of the way through the download, Windows forces a restart to install updates, which takes 20 minutes. Then Carla has to restart the download from the beginning. How long does it take to download the file? Reference:} \texttt{<<200*40*.01=80>>}, \texttt{<<80/2=40>>},
\texttt{<<200/2=100>>}, \texttt{<<40+100+20=160>>}. \emph{Gold answer $160$.}

\begin{quote}\small
\textbf{(A) \modelName{} LM head} ($\loopHidden{\numLoopPasses}$ through $\lmhead$).
\begin{Verbatim}[commandchars=\@\[\]]
B0: @rdhl[<<200/2=6=120>>]\n6>>\n>>\n>>\n>>\n>>\n2>>\n505050>>\n605050<...
B1: @rdhl[<<40*.2=40>>]\n120>>\n666>>\n>>\n>>\n>>\n>>\n60>>\n@rdhl[80]80>>\n>>...
B2: @rdhl[<<60+40=120>>]\n8>>\n>>\n>>\n>>\n>>\n>>\n>>\n6050>>\n>>\n>>\n...
B3: @rdhl[<<140+20=120>>]\n8>>\n33>>\n>>\n>>\n>>\n>>\n2020=120>>\n>>\n>...
B4: @rdhl[<<140+20=120>>]\n120>>\n33>>\n>>\n>>\n>>\n###+60+/1>>\n>>\n>>...
B5: @rdhl[160]160+@rdhl[20]=120>>\n180>>\n>>\n>>\n>>\n180>>\n###+600+100120>>\...
\end{Verbatim}

\textbf{(B) \modelName-aux LM head} ($\loopHidden{\numLoopPasses}$ through $\auxhead$).
\begin{Verbatim}[commandchars=\@\[\]]
B0: 100200/) @rdhl[100]100))<<100100**@rdhl[40] percentage,404040,<<100-+
B1: @rdhl[20]20= @rdhl[80]+ and###@rdhl[100]100+20++ 120120)###100100+220+
B2:  120120 plus###@rdhl[100]100*10020 140 plus###12080+2100= 120240)\n...
B3: ###120120+@rdhl[20]20+120140140######120+202020 140140)\n######120+
B4: +2=  140140)\n)\n######120120+@rdhl[20]2020  140100)\n Ris######
B5: ###120+@rdhl[20]20= @rdhl[100]100 Ris#########120+20= 100100 Ris##########...
\end{Verbatim}

\textbf{(C) \modelName-aux auxiliary decoder} ($\auxHidden{t}$ through $\auxhead$, FR).\\
\texttt{\rdhl{<<100+20=120>>}\textbackslash n\rdhl{<<120+20=140>>}\textbackslash n\rdhl{<<100+20=120>>}\textbackslash n\#\#\# 140}

\textbf{Final answer (autoregressive, after $6$ loops).} \modelName{}: \texttt{\#\#\# 180}. \modelName-aux: \texttt{\#\#\# 120}. Gold: $160$.

\textbf{Reading the chains.} The latents recover most of the right intermediates ($\mathbf{80, 100, 40, 20}$) but fail to combine them: (A) flickers between $120, 140, 160, 180$ in the late blocks; (B) commits to $120 + 20$ chains; (C) outputs \texttt{<<100+20=120>>, <<120+20=140>>, ...} and emits $140$. Both methods land on plausible-but-wrong totals.
\end{quote}

\paragraph{Example 4 — Melanie's vacuums (failure).} \quad

\emph{Question: Melanie is a door-to-door saleswoman. She sold a third of her vacuum cleaners at the green house, 2 more to the red house, and half of what was left at the orange house. If Melanie has 5 vacuum cleaners left, how many did she start with? Reference:} \texttt{<<5*2=10>>}, \texttt{<<10+2=12>>}. \emph{Gold answer $18$. One extra step is omitted from the reference; the true chain is $5\!\to\!10\!\to\!12\!\to\!18$ ($12 \times 3/2 = 18$).}

\begin{quote}\small
\textbf{(A) \modelName{} LM head} ($\loopHidden{\numLoopPasses}$ through $\lmhead$).
\begin{Verbatim}[commandchars=\@\[\]]
B0: @rdhl[<<5*2=10>>]\n33>>\n@rdhl[12]>>\n12>>\n22=1312>>\n1212121313<<
B1: @rdhl[<<10+2=17>>]\n@rdhl[5]>>\n5>>\n17>>\n>>\n1//1181818.55<<
B2: @rdhl[<<9/(1/2/5=20.5>>]\n20>>\n15.18>>\n>>\n>>\n>>\n5>>\n<<
B3: @rdhl[<<15*1=30>>]\n15>>\n@rdhl[5]>>\n5>>\n20>>\n5=151818>>\n>>\n55<<
B4: @rdhl[<<3=15>>]\n30>>\n15>>\n@rdhl[5]>>\n5>>\n>>\n>>\n55515>>\n15>>\n5>>\n...
B5: @rdhl[<<15+5=15>>]\n15>>\n5>>\n>>\n>>\n17>>\n55515>>\n@rdhl[18]>>\n18>>\n>...
\end{Verbatim}

\textbf{(B) \modelName-aux LM head} ($\loopHidden{\numLoopPasses}$ through $\auxhead$).
\begin{Verbatim}[commandchars=\@\[\]]
B0: 105* half @rdhl[10]10)+101010+2)= @rdhl[12]12)<<1212*3
B1: 3)1512..<<@rdhl[12]1212*3 third = 3636.###121212*3)
B2:   3636)######@rdhl[12]+33= 3636)######36+125= 
B3: 3636/#########36+@rdhl[12]++  4836 Egg#########36+12+= 
B4: 483636#########+@rdhl[5]== 4836############36+5  3636###
B5: ############15+3  3636###############15###1 = 363615###
\end{Verbatim}

\textbf{(C) \modelName-aux auxiliary decoder} ($\auxHidden{t}$ through $\auxhead$, FR).\\
\texttt{\rdhl{<<10+2=12>>}\textbackslash n\rdhl{<<10+2=12>>}\textbackslash n\rdhl{<<10+2=12>>}\textbackslash n\rdhl{<<12+2=14>>}\textbackslash n...}

\textbf{Final answer (autoregressive, after $6$ loops).} \modelName{}: \texttt{\#\#\# 30}. \modelName-aux: \texttt{\#\#\# 36}. Gold: $18$.

\textbf{Reading the chains.} (A): \modelName{}'s latents flicker between $15, 18, 30$ across blocks. The valid sub-chain $\mathbf{5\!\to\!10\!\to\!12\!\to\!18}$ surfaces in B5, but the final readout commits to $30$. (B): \modelName-aux locks onto a wrong $\times 3$ branch (presumably ``a third'' read as multiply-by-three), with $36$ saturating B4 to B5. (C): the aux decoder loops on \texttt{<<10+2=12>>}.
\end{quote}

\textbf{What these readouts are (and are not).} The readouts (A), (B), (C) are just diagnostic projections of the post-loop continuous latents back into the discrete token space. They are not part of the inference pipeline. At inference, the final answer is generated autoregressively from the continuous post-loop latents (the ``Final answer'' line in each example), conditioning on the latent representation itself rather than on any discretized chain. The intermediate streams are intended to give intuition about \emph{what numbers the latents place mass on}, not as a quality measure of the model's reasoning: a noisy or repetitive (A)/(B) stream is compatible with a correct final answer, and a clean (C) chain is not required for the model to be correct.

\subsection{Details for Multi-Path Analysis\texorpdfstring{~(\cref{sec:exp:analysis:multipath})}{}}
\label{app:multipath-setup}
\paragraph{Datasets.} Note that GSM8K-Aug~\citep{deng2023implicit} augments the original GSM8K~\citep{cobbe2021trainingverifierssolvemath} training split (about $7.5$K problems) into the $385{,}620$ training examples we use. 
GSM8K is thus a subset of GSM8K-Aug, and the multi-path bank below is built from the \emph{original} GSM8K problems, which are also in the training set.
\begin{figure}[t]
    \centering
    \includegraphics[width=0.9\linewidth]{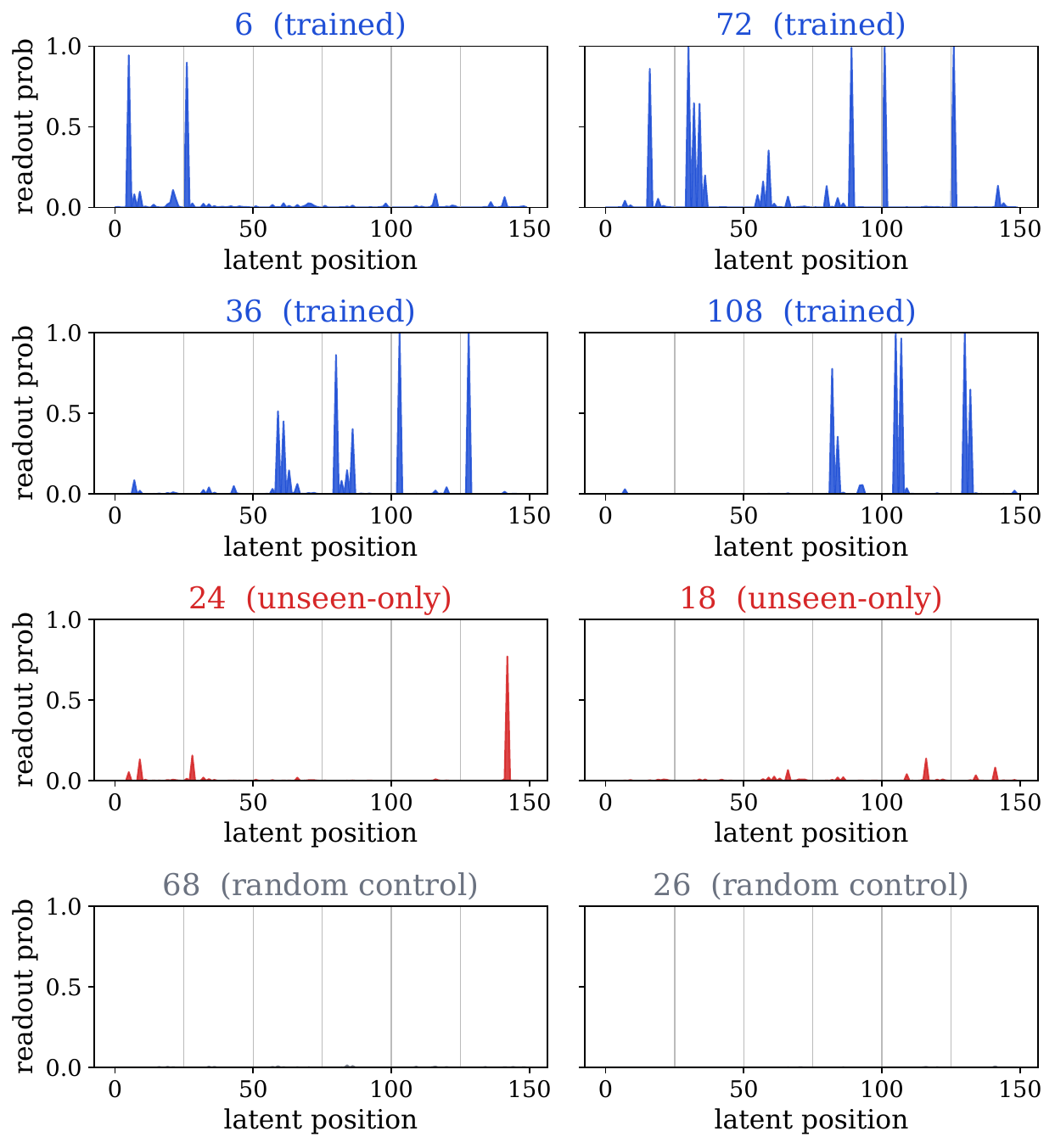}
    \caption{{Multi-path readout probability for the example in \cref{app:multipath-example}. The only-$U$ intermediates do not appear in the trained path, question, or answer, but they surface in the post-loop latent readout.}}
    \label{fig:multipath-grid6}
\end{figure}
\paragraph{Path bank.} For each question we identify its trained path (its gold chain in the training data) and an unseen-but-valid alternative drawn from a bank of up to ten correct paths obtained by rejection-sampling \textsc{Llama-3.1-8B}~\citep{grattafiori2024llama} on the original GSM8K training questions ($7{,}444$ questions with $58{,}155$ verified paths in total), with each alternative verified absent from training.

\paragraph{Candidate sets and filtering.} We partition intermediate numbers into three disjoint sets: $G$ (trained-path only), $U$ (unseen-path only), and a random control from other questions matched to $U$ in both cardinality and single/multi-token composition. Numbers shared by both paths or appearing in the question or answer are excluded, since they cannot indicate which chain a readout recovered. The random control is fixed and shared across all four readouts and verified disjoint from the trained path, unseen path, question, and answer numbers. Intersecting the bank with our training set and keeping questions with $\geq 2$ disjoint intermediates per side yields {$653$ candidate questions. We then apply a token-clean filter, requiring every only-$U$ number to share no sub-token with any trained-path, question, or answer number, leaving the final $341$ questions, spanning $871$ $G$, $974$ $U$, and $974$ Random numbers. Random stays matched to $U$ in cardinality and composition under every criterion.} 

\paragraph{Metrics.} We score a target number by its best-position, per-token NLL. A number spans one or more consecutive sub-tokens, and at every latent position the readout gives a softmax distribution over the vocabulary. We slide the number's sub-tokens across the latent positions. At each placement we average the sub-tokens' NLLs at their aligned positions, and keep the lowest such per-token average over all placements (for a single-token number, simply the negative log of its highest readout probability across positions). Per-token normalization puts single and multi-token numbers on the same scale, and scoring only the most probable span avoids double counting overlaps. We average this over the numbers in sets $G$, $U$, and Random within each question, then over the {$341$} questions. The $P(U\in\text{top-}k)$ columns instead report a hit rate: the fraction of $U$ numbers recovered within the top-$k$ candidates (top-$k$ membership at any position for single-token numbers, and a consecutive top-$k$ match for multi-token ones), counting each number once however many positions it surfaces at, averaged within each question and then over the {$341$} questions.

\subsection{Multi-Path Readout Example}
\label{app:multipath-example}
{This example (referenced from \cref{sec:exp:analysis:multipath}) shows a question that admits two valid reasoning chains with the same answer $108$, read out by \modelName{} ($\loopHidden{\numLoopPasses}$ through the base LM head $\lmhead$, i.e.\ the first row of \cref{tab:path-mass}).}
\begin{quote}\small
{\textit{Q:} The bus driver drives an average of 2 hours each day, 5 days a week. From Monday to Wednesday he drove at an average speed of 12 kilometers per hour, and from Thursday to Friday at an average speed of 9 kilometers per hour. How many kilometers did the driver travel during these 5 days?}\\[3pt]
{\textit{Trained path $G$:} \texttt{3*2=6 -> 6*12=72 -> 2*2=4 -> 4*9=36 -> 72+36=108}.}\\
{\textit{Unseen path $U$:} \texttt{2*12=24 -> 24+24+24=72 -> 2*9=18 -> 18+18=36 -> 72+36=108}.}\\
{$\text{only}_G = \{3, 4, 6\}$,\quad $\text{only}_U = \{24, 18\}$\quad (question and final answer excluded).}
\end{quote}
{\Cref{fig:multipath-grid6} visualizes the same example, aligning the trained path, the unseen valid path, and the latent readout positions where the only-$U$ intermediates surface.} {The only-$U$ intermediates $24$ and $18$ are the per-day distances in the alternative decomposition, which forms daily distances before summing rather than aggregating hours first. Although neither appears in the trained reference path, both surface in the post-loop latent readout through the base LM head: $P(\texttt{24}){=}0.77$ (top-$1$, position $142$) and $P(\texttt{18}){=}0.14$ (top-$2$, position $116$). Every number here is a single token, so no only-$U$ number shares a sub-token with any trained-path, question, or answer number, confirming the surfacing is genuine unseen-chain content rather than a digit-overlap artifact.}

%% file: related_appendix.tex
\section{Related Work}
\label{app:related}

We refer the reader to \citet{chen2025reasoninglanguagecomprehensivesurvey,yu2026latentspacefoundationevolution,zhu2025surveylatentreasoning,li2025implicitreasoninglargelanguage,feng2025efficientreasoningmodelssurvey}, and \citet{maile2026dynamicparameterreuse} for broad surveys of the latent reasoning and looped transformers landscapes. Below, we discuss the work most directly adjacent to ours.

\paragraph{Looped and recurrent-depth transformers.}
Building on Universal Transformers \citep{dehghani2019universaltransformers}, a line of work has established the theoretical foundations of looped architectures: \citet{pmlr-v202-giannou23a} show Turing completeness via program simulation, placing looped transformers alongside CoT in expressivity \citep{merrill2024the,li2024chain,nowak-etal-2024-representational}.
\citet{saunshi2025reasoninglatentthoughtspower,merrill2025littledepthgoeslong,merrill2025exactexpressivepowertransformers,jerad2026contextfreerecognitiontransformers} study the efficient reasoning abilities of looping, establishing their ability to solve problems with fewer model evaluations than CoT.
\citet{saunshi2025reasoninglatentthoughtspower} show that a shallow Transformer looped multiple times nearly matches a non-looped Transformer with the equivalent total depth on arithmetic and reasoning tasks, with looping implicitly simulating hidden reasoning steps.
\citet{xu2025cotloopformalcomparison,Xu2025AFC} compare looped Transformers to CoT, and \citet{svete2026on} draw connections to masked diffusion models.
The latter connection also inspires approaches to improve looped transformers with time modulation \citep{jeddi2026loopformer,chen2026thinkingdeeperlongerdepthrecurrent} and efficient sampling \citep{geiping2025efficient}.
An active line of work scales looped models to large-scale pretraining: \citet{geiping2025scalingup} train a 3.5B recurrent-depth Transformer whose loop depth can be increased at test time, \citet{zhu2025scalinglatentreasoninglooped} pretrain LoopLM with token-level recurrence and learned depth allocation, and \citet{zeng2026ponderlm,zeng2026ponderlm2pretrainingllmlatent} pretrain models that loop by superpositioning symbols in continuous space based on the logits produced after each iteration. 
LoopLM shows that looped computation can scale in pretraining, but its per-token loop at inference time does not by itself create the parallel padded latent workspace that \modelName{} uses.
Complementary work addresses different ways of implementing \emph{adaptive compute} with looped architectures: Depth can be modulated per-symbol via gating \citep{ng2024loopneuralnetworksparameter,song2026adaponderlmgatedponderinglanguage,moosa2026understandingdynamiccomputeallocation}, mixture-of-depths routing \citep{raposo2024mixtureofdepthsdynamicallyallocatingcompute}, or RL-trained stopping criteria \citep{ning2025learningstopadaptivelatent}.
Depth-recurrent attention mixtures augment recurrence with attention mechanisms that can attend to latent representations of past iterations \citep{fu2025thinkathardselectivelatentiterations,knupp2026depthrecurrentattentionmixturesgiving}.
Parameter efficiency and flexibility are addressed by per-iteration LoRA adapters \citep{bae2025relaxed}, symbol-level recursion routing \citep{bae2025mixtureofrecursions}, inner thinking mechanisms \citep{chen-etal-2025-inner}, and Mixture-of-Experts compatibility \citep{csordas2024moeut}.
Further work explores retrofitting recurrence into pretrained models \citep{mcleish2025teachingpretrainedlanguagemodels,koishekenov2025encodethinkdecodescaling}, compositional and length generalization \citep{fan2025looped,kohli2026loopthinkgeneralize}, reinforcement pretraining \citep{tang2026looprptreinforcementpretraininglooped}, and whether pretrained models use their depth efficiently \citep{csordas2025do}.

\paragraph{Diffusion language models.}
Diffusion language models are another route to parallel, non-autoregressive reasoning. Discrete-token variants face a \emph{parallel-decoding trap}, where tokens denoised in parallel are conditionally independent, so recovering quality forces a partial return to sequential decoding \citep{liu2025tidar}. Notably, \citet{svete2026on} show masked diffusion models are equivalent to padded looped Transformers, motivating supervising such a backbone directly, as we do. Continuous diffusion variants such as LaDiR~\citep{kang2026ladir} instead refine latent thoughts with a VAE and a separate diffusion model over many denoising steps for diversity and adaptive compute, while showing limited efficiency gain when reaching \CoTAcr-level performance.

\paragraph{Pause and padding symbols.}
A parallel strand of work studies the affordances of discrete padding symbols.
\citet{goyal2024think} and \citet{pfau2024lets} show that they improve performance on natural language and formal tasks; \citet{wang2024guiding} use discrete symbols in the continuous representation space as planning signals; and \citet{chu2026spotspanlevelpauseofthoughtefficient} introduce span-level pause symbols for efficient and interpretable latent reasoning.
These discrete-symbol approaches motivate our use of continuous latent vectors as a richer computation medium, while our looped architecture lets each iteration refine the same latent rather than extending the sequence.

\paragraph{Compressing and replacing discrete CoT.}
Latent CoT methods replace or compress explicit reasoning symbols with continuous representations.
\citet{cheng2024compressedchainthoughtefficient} supervise hidden-state representations of salient reasoning steps, \citet{tan2025think} dynamically compress CoT chains into latent thoughts, \citet{wang2026latentchainofthoughtplanningdecoupling} autoregressively encode CoT steps into hidden states decoded back to ground-truth symbols (analogously to our approach), and \citet{amos2026latentreasoningsupervisedthinking} interleave supervised latent states with generated text.
MUX~\citep{suleymanzade2026mux} studies continuous reasoning with multiplexed vocabulary-space targets. It uses KL regression to multiplexed targets. Each target is a position-weighted superposition over aligned reasoning subwords, so several subwords are mixed into one sequential latent. In contrast, \modelName{} keeps step and token positions explicit in a parallel padded workspace and applies per-position cross-entropy on gold \CoTAcr traces.
Latent CoT also enables parallel exploration of reasoning paths \citep{gozeten2026continuous,liu2026thoughtbubblesunsupervisedmethodparallel}, and test-time compute can be scaled in latent space via policy optimization \citep{ye2026thinking}, instance-level policy gradient \citep{li2026seekdarkreasoningtesttime}, sampling conditioned on perturbed prompts \citep{xu2025softcottesttimescalingsoft}, or parallel search \citep{lian2025threadweaveradaptivethreadingefficient,you2026paralleltesttimescalinglatent}.
Latent reasoning can be integrated with generation by mixing latent and text symbols \citep{fu2026selarselectivelatentreasoning,deng2026llmlatentreasoningchain}, updating the KV cache with an additional model \citep{liu2025deliberation}, learning meta-symbols that guide generation \citep{shah2025language}, and semantically aligning padding symbols to spans of CoT symbols \citep{he2025semcot}.
Further work studies latent reasoning with variational autoencoder features \citep{su2025token,kang2026ladir} and energy-based guidance \citep{chen2025thinkconsistentlyreasonefficiently}. GRAM~\citep{baek2026generative} guides thinking with recursive latent-variable modeling through shared transitions. It uses stochastic high-level updates and learns a target-conditioned posterior and prior through amortized variational inference. GRAM samples distinct stochastic trajectories, whereas \modelName{} represents them within a jointly refined continuous workspace. Interpretability studies probe whether latents actually contribute to reasoning \citep{zhang2025latenttokensthinkcausal,rizvi-martel2026the}, and empirical comparisons examine the effects of different supervision regimes \citep{cui2026latentreasoningmethodsperform,suleymanzade2026limits}.

\paragraph{Speculative decoding.} Speculative decoding uses draft models to propose token continuations \citep{leviathan2023fast,chen2023accelerating}. Related approaches include Lookahead~\citep{fu2024lookahead} with Jacobi-style updates, Medusa~\citep{cai2024medusa} with auxiliary prediction heads, and EAGLE~\citep{li2024eagle} with future hidden features. These methods parallelize proposals within an explicit token trace, whereas \modelName{} parallelizes latent reasoning and decodes only the answer suffix.

%% file: notation.tex

\begin{table}[t]
  \centering
  \small
  \caption{Notation used in this paper.}
  \label{tab:notation}
  \begin{tabular}{@{}lll@{}}
    \toprule
    Symbol & Meaning & First used \\
    \midrule
    \multicolumn{3}{l}{\emph{Inputs and targets}} \\
    $\instr$                       & Question                                                   & \S\ref{sec:prelim} \\
    $\outstr$                      & Answer suffix (ends with $\langle\text{EoS}\rangle$)        & \S\ref{sec:method:prefix} \\
    $\stepTokens{\blockidx}$       & Tokenized gold \CoTAcr step $\blockidx$ ($T_{\blockidx,j}$ its $j$-th token) & \S\ref{sec:prelim} \\
    $\thoughtSeq$                  & Full gold \CoTAcr token sequence $(\stepTokens{1},\dots,\stepTokens{\blockBudget})$ & \S\ref{sec:method:supervision} \\
    \midrule
    \multicolumn{3}{l}{\emph{Counts and budgets (fixed vs.\ example-dependent)}} \\
    $\numCoTSteps$                 & Number of \CoTAcr steps in an example (\emph{example-dependent})         & \S\ref{sec:prelim} \\
    $\numCoTTokens$                & Total number of \CoTAcr tokens in an example                            & \S\ref{sec:prelim} \\
    $\blockBudget$                 & Latent block budget (\emph{fixed}; chosen so $\blockBudget\ge\numCoTSteps$) & \S\ref{sec:method:prefix} \\
    $\tokPerBlock$                 & Latent tokens per block (block width)                                    & \S\ref{sec:method:prefix} \\
    $\numLoopPasses$               & Number of loop iterations (depth)                                        & \S\ref{sec:method:loop} \\
    $\embedDimLM$                  & Hidden / embedding dimension of the base LM                             & \S\ref{sec:method:loop} \\
    \midrule
    \multicolumn{3}{l}{\emph{Indices}} \\
    $\blockidx$                    & Latent block index, $\blockidx\in\{1,\dots,\blockBudget\}$              & \S\ref{sec:method:prefix} \\
    $\loopPassIdx$                 & Loop pass index, $\loopPassIdx\in\{1,\dots,\numLoopPasses\}$            & \S\ref{sec:method:loop} \\
    {$j$}                     & {Token index within a \CoTAcr step / latent block}                & \S\ref{sec:prelim} \\
    \midrule
    \multicolumn{3}{l}{\emph{Model and special tokens}} \\
    $\baseLM$                      & Base causal LM (parameters $\params$)                                   & \S\ref{sec:method:loop} \\
    $\lmhead$                      & Base LM head (logits from a hidden state)                               & \S\ref{sec:method:supervision} \\
    $g_\phi$                       & Auxiliary decoder (\modelName-aux, training only); head $\auxhead$ & \S\ref{sec:method:lotus-aux} \\
    $\auxhead$                     & {Auxiliary decoder LM head}                                        & \S\ref{sec:method:lotus-aux} \\
    $\latTokenSym$                 & Learnable latent token (shared across positions)                        & \S\ref{sec:method:prefix} \\
    $\latRegionStart,\,\latRegionEnd$ & Begin-/end-of-thought delimiters                                     & \S\ref{sec:method:prefix} \\
    $\cache_\text{pre}$            & Prefix KV cache (computed once, reused)                                 & \S\ref{sec:method:loop} \\
    $\loopEmb$                     & Learnable latent embeddings fed to the loop                             & \S\ref{sec:method:loop} \\
    \midrule
    \multicolumn{3}{l}{\emph{Hidden states (distinct symbols by source)}} \\
    $\loopHidden{\loopPassIdx}$    & Post-loop latent states after pass $\loopPassIdx$ (entries $\loopHidden[\blockidx,j]{\loopPassIdx}$) & \S\ref{sec:method:loop} \\
    $\ansHidden$                   & Final-forward hidden states for the answer loss                         & \S\ref{sec:method:supervision} \\
    $\auxHidden{\loopPassIdx}$     & Auxiliary-decoder hidden states at iteration $\loopPassIdx$             & \S\ref{sec:method:lotus-aux} \\
    \midrule
    \multicolumn{3}{l}{\emph{Losses and objective}} \\
    $\stepLoss$                    & Parallel step supervision loss (\cref{eq:step})                         & \S\ref{sec:method:supervision} \\
    $\ansLoss$                     & Answer-suffix next-token loss                                           & \S\ref{sec:method:supervision} \\
    $\totalLoss$                   & Full objective, $\ansLoss+\stepLossWeight\,\stepLoss$                   & \S\ref{sec:method:supervision} \\
    $\stepLossWeight$              & Weight on the step loss                                                 & \S\ref{sec:method:supervision} \\
    $\normStepToks$                & Number of supervised (non-padding) \CoTAcr tokens                       & \S\ref{sec:method:supervision} \\
    $\CE(\cdot,\cdot)$             & Cross-entropy                                                      & \S\ref{sec:method:supervision} \\
    $\auxStepLoss$                 & {Auxiliary-decoder step loss (\cref{eq:aux})}                    & \S\ref{sec:method:lotus-aux} \\
    $\auxLoss$                     & {\modelName-aux objective, $\ansLoss+\stepLossWeight\,\auxStepLoss$} & \S\ref{sec:method:lotus-aux} \\
    \midrule
    \multicolumn{3}{l}{\emph{Analysis (\cref{sec:method:theory})}} \\
    $\questionVar$                 & Question random variable                                                & \S\ref{sec:method:theory} \\
    $\dataChainDist$               & True data distribution over chains                                      & \S\ref{sec:method:theory} \\
    $p_\params^{\mathrm{PCL}}$     & Parallel chain likelihood induced by $\stepLoss$                        & \S\ref{sec:method:theory} \\
    \bottomrule
  \end{tabular}
\end{table}

%% file: references.bib
@inproceedings{li2024eagle,
  title={EAGLE: Speculative Sampling Requires Rethinking Feature Uncertainty},
  author={Li, Yuhui and Wei, Fangyun and Zhang, Chao and Zhang, Hongyang},
  booktitle={International Conference on Machine Learning},
  pages={28935--28948},
  year={2024},
  organization={PMLR}
}

@inproceedings{cai2024medusa,
  title={Medusa: Simple LLM Inference Acceleration Framework with Multiple Decoding Heads},
  author={Cai, Tianle and Li, Yuhong and Geng, Zhengyang and Peng, Hongwu and Lee, Jason D and Chen, Deming and Dao, Tri},
  booktitle={International Conference on Machine Learning},
  pages={5209--5235},
  year={2024},
  organization={PMLR}
}

@inproceedings{fu2024lookahead,
  title={Break the Sequential Dependency of LLM Inference Using Lookahead Decoding},
  author={Fu, Yichao and Bailis, Peter and Stoica, Ion and Zhang, Hao},
  booktitle={International Conference on Machine Learning},
  pages={14060--14079},
  year={2024},
  organization={PMLR}
}

@article{chen2023accelerating,
  title={Accelerating large language model decoding with speculative sampling},
  author={Chen, Charlie and Borgeaud, Sebastian and Irving, Geoffrey and Lespiau, Jean-Baptiste and Sifre, Laurent and Jumper, John},
  journal={arXiv preprint arXiv:2302.01318},
  year={2023}
}

@inproceedings{leviathan2023fast,
  title={Fast inference from transformers via speculative decoding},
  author={Leviathan, Yaniv and Kalman, Matan and Matias, Yossi},
  booktitle={International Conference on Machine Learning},
  pages={19274--19286},
  year={2023},
  organization={PMLR}
}

@article{baek2026generative,
  title={Generative Recursive Reasoning},
  author={Baek, Junyeob and Jo, Mingyu and Kim, Minsu and Ren, Mengye and Bengio, Yoshua and Ahn, Sungjin},
  journal={arXiv preprint arXiv:2605.19376},
  year={2026}
}

@article{deng2023implicit,
  title={Implicit chain of thought reasoning via knowledge distillation},
  author={Deng, Yuntian and Prasad, Kiran and Fernandez, Roland and Smolensky, Paul and Chaudhary, Vishrav and Shieber, Stuart},
  journal={arXiv preprint arXiv:2311.01460},
  year={2023}
}

@inproceedings{
suleymanzade2026mux,
title={{MUX}: Continuous Reasoning via Multiplexed Tokens},
author={Ayhan Suleymanzade and Halil Alperen Gozeten and Ismail Ilkan Ceylan and Jinwoo Kim},
booktitle={ICLR 2026 Workshop on Logical Reasoning of Large Language Models},
year={2026},
url={https://openreview.net/forum?id=Ee78Uqsq0a}
}

@inproceedings{
kuzina2026kava,
title={KaVa: Latent Reasoning via Compressed {KV}-Cache Distillation},
author={Anna Kuzina and Maciej Pi{\'o}ro and Babak Ehteshami Bejnordi},
booktitle={The Fourteenth International Conference on Learning Representations},
year={2026},
url={https://openreview.net/forum?id=ePrhcLbtGv}
}

@inproceedings{jeddi2026loopformer,
  author    = {Ahmadreza Jeddi and Marco Ciccone and Babak Taati},
  booktitle = {The Fourteenth International Conference on Learning Representations},
  title     = {LoopFormer: Elastic-Depth Looped Transformers for Latent Reasoning via Shortcut Modulation},
  url       = {https://openreview.net/forum?id=RzYXb5YWBs},
  year      = {2026}
}

@article{ning2025learningstopadaptivelatent,
  archiveprefix = {arXiv},
  author        = {Alex Ning and Yen-Ling Kuo and Gabe Gomes},
  eprint        = {2511.21581},
  primaryclass  = {cs.LG},
  title         = {Learning When to Stop: Adaptive Latent Reasoning via Reinforcement Learning},
  url           = {https://arxiv.org/abs/2511.21581},
  year          = {2025}
}

@inproceedings{shah2025language,
  author    = {Alok Shah and Khush Gupta and Keshav Ramji and Pratik Chaudhari},
  booktitle = {ICML 2025 Workshop on Long-Context Foundation Models},
  title     = {Language Modeling with Learned Meta-Tokens},
  url       = {https://openreview.net/forum?id=oaHYnLldHM},
  year      = {2025}
}

@inproceedings{svete2026on,
  author    = {Anej Svete and Ashish Sabharwal},
  booktitle = {The Fourteenth International Conference on Learning Representations},
  title     = {On the Reasoning Abilities of Masked Diffusion Language Models},
  url       = {https://openreview.net/forum?id=BVnIsh4Nz1},
  year      = {2026}
}

@article{altabaa2025unlockingoutofdistributiongeneralizationtransformers,
  archiveprefix = {arXiv},
  author        = {Awni Altabaa and Siyu Chen and John Lafferty and Zhuoran Yang},
  eprint        = {2510.14095},
  journal       = {arXiv preprint arXiv:2510.14095},
  primaryclass  = {cs.LG},
  title         = {Unlocking Out-of-Distribution Generalization in Transformers via Recursive Latent Space Reasoning},
  url           = {https://arxiv.org/abs/2510.14095},
  year          = {2025}
}

@inproceedings{suleymanzade2026limits,
  author    = {Ayhan Suleymanzade and Andreas Bergmeister and Stefanie Jegelka},
  booktitle = {Logical and Symbolic Reasoning in Language Models @ AAAI 2026},
  title     = {Limits of Continuous Chain-of-Thought in Multi-Step and Multi-Chain Reasoning},
  url       = {https://openreview.net/forum?id=UQFTJPqJAc},
  year      = {2026}
}

@article{zeng2026ponderlm2pretrainingllmlatent,
  archiveprefix = {arXiv},
  author        = {Boyi Zeng and He Li and Shixiang Song and Yixuan Wang and Zitong Wang and Ziwei He and Xinbing Wang and Zhouhan Lin},
  eprint        = {2509.23184},
  journal       = {arXiv preprint arXiv:2509.23184},
  primaryclass  = {cs.CL},
  title         = {PonderLM-2: Pretraining LLM with Latent Thoughts in Continuous Space},
  url           = {https://arxiv.org/abs/2509.23184},
  year          = {2026}
}

@inproceedings{zeng2026ponderlm,
  author    = {Boyi Zeng and Shixiang Song and Siyuan Huang and Yixuan Wang and He Li and Ziwei He and Xinbing Wang and Zhiyu li and Zhouhan Lin},
  booktitle = {The Fourteenth International Conference on Learning Representations},
  title     = {Ponder{LM}: Pretraining Language Models to Ponder in Continuous Space},
  url       = {https://openreview.net/forum?id=UrM4MNRYZm},
  year      = {2026}
}

@inproceedings{chen-etal-2025-inner,
  address   = {Vienna, Austria},
  author    = {Chen, Yilong  and
               Shang, Junyuan  and
               Zhang, Zhenyu  and
               Xie, Yanxi  and
               Sheng, Jiawei  and
               Liu, Tingwen  and
               Wang, Shuohuan  and
               Sun, Yu  and
               Wu, Hua  and
               Wang, Haifeng},
  booktitle = {Proceedings of the 63rd Annual Meeting of the Association for Computational Linguistics (Volume 1: Long Papers)},
  doi       = {10.18653/v1/2025.acl-long.1369},
  editor    = {Che, Wanxiang  and
               Nabende, Joyce  and
               Shutova, Ekaterina  and
               Pilehvar, Mohammad Taher},
  isbn      = {979-8-89176-251-0},
  month     = jul,
  pages     = {28241--28259},
  publisher = {Association for Computational Linguistics},
  title     = {Inner Thinking Transformer: Leveraging Dynamic Depth Scaling to Foster Adaptive Internal Thinking},
  url       = {https://aclanthology.org/2025.acl-long.1369/},
  year      = {2025}
}

@article{raposo2024mixtureofdepthsdynamicallyallocatingcompute,
  archiveprefix = {arXiv},
  author        = {David Raposo and Sam Ritter and Blake Richards and Timothy Lillicrap and Peter Conway Humphreys and Adam Santoro},
  eprint        = {2404.02258},
  journal       = {arXiv preprint arXiv:2404.02258},
  primaryclass  = {cs.LG},
  title         = {Mixture-of-Depths: Dynamically allocating compute in transformer-based language models},
  url           = {https://arxiv.org/abs/2404.02258},
  year          = {2024}
}

@article{deepseek-r1,
  author   = {DeepSeek-AI},
  day      = {01},
  doi      = {10.1038/s41586-025-09422-z},
  issn     = {1476-4687},
  journal  = {Nature},
  month    = {Sep},
  number   = {8081},
  pages    = {633-638},
  title    = {{DeepSeek-R1} incentivizes reasoning in {LLMs} through reinforcement learning},
  url      = {https://doi.org/10.1038/s41586-025-09422-z},
  volume   = {645},
  year     = {2025}
}

@inproceedings{su2025token,
  author    = {DiJia Su and Hanlin Zhu and Yingchen Xu and Jiantao Jiao and Yuandong Tian and Qinqing Zheng},
  booktitle = {Forty-second International Conference on Machine Learning},
  title     = {Token Assorted: Mixing Latent and Text Tokens for Improved Language Model Reasoning},
  url       = {https://openreview.net/forum?id=hYfOPXrbUr},
  year      = {2025}
}

@inproceedings{pmlr-v202-giannou23a,
  author    = {Giannou, Angeliki and Rajput, Shashank and Sohn, Jy-Yong and Lee, Kangwook and Lee, Jason D. and Papailiopoulos, Dimitris},
  booktitle = {Proceedings of the 40th International Conference on Machine Learning},
  editor    = {Krause, Andreas and Brunskill, Emma and Cho, Kyunghyun and Engelhardt, Barbara and Sabato, Sivan and Scarlett, Jonathan},
  month     = {23--29 Jul},
  pages     = {11398--11442},
  publisher = {PMLR},
  series    = {Proceedings of Machine Learning Research},
  title     = {Looped Transformers as Programmable Computers},
  url       = {https://proceedings.mlr.press/v202/giannou23a.html},
  volume    = {202},
  year      = {2023}
}

@article{tang2026looprptreinforcementpretraininglooped,
  archiveprefix = {arXiv},
  author        = {Guo Tang and Shixin Jiang and Heng Chang and Nuo Chen and Yuhan Li and Huiming Fan and Jia Li and Ming Liu and Bing Qin},
  eprint        = {2603.19714},
  journal       = {arXiv preprint arXiv:2603.19714},
  primaryclass  = {cs.CL},
  title         = {LoopRPT: Reinforcement Pre-Training for Looped Language Models},
  url           = {https://arxiv.org/abs/2603.19714},
  year          = {2026}
}

@inproceedings{gozeten2026continuous,
  author    = {Halil Alperen Gozeten and Muhammed Emrullah Ildiz and Xuechen Zhang and Hrayr Harutyunyan and Ankit Singh Rawat and Samet Oymak},
  booktitle = {The Fourteenth International Conference on Learning Representations},
  title     = {Continuous Chain of Thought Enables Parallel Exploration and Reasoning},
  url       = {https://openreview.net/forum?id=sTPKDKn5ig},
  year      = {2026}
}

@inproceedings{zhu2025reasoning,
  author    = {Hanlin Zhu and Shibo Hao and Zhiting Hu and Jiantao Jiao and Stuart Russell and Yuandong Tian},
  booktitle = {ICML 2025 Workshop on Methods and Opportunities at Small Scale},
  title     = {Reasoning by Superposition: A Theoretical Perspective on Chain of Continuous Thought},
  url       = {https://openreview.net/forum?id=1cD9iO5Isv},
  year      = {2025}
}

@inproceedings{kang2026ladir,
  author    = {Haoqiang Kang and Yizhe Zhang and Nikki Lijing Kuang and Nicklas Majamaki and Navdeep Jaitly and Yian Ma and Lianhui Qin},
  booktitle = {The Fourteenth International Conference on Learning Representations},
  title     = {LaDiR: Latent Diffusion Enhances {LLM}s for Text Reasoning},
  url       = {https://openreview.net/forum?id=z5cPEZ4n6i},
  year      = {2026}
}

@article{kohli2026loopthinkgeneralize,
  archiveprefix = {arXiv},
  author        = {Harsh Kohli and Srinivasan Parthasarathy and Huan Sun and Yuekun Yao},
  eprint        = {2604.07822},
  journal       = {arXiv preprint arXiv:2604.07822},
  primaryclass  = {cs.CL},
  title         = {Loop, Think, \& Generalize: Implicit Reasoning in Recurrent-Depth Transformers},
  url           = {https://arxiv.org/abs/2604.07822},
  year          = {2026}
}

@article{li2026seekdarkreasoningtesttime,
  archiveprefix = {arXiv},
  author        = {Hengli Li and Chenxi Li and Tong Wu and Xuekai Zhu and Yuxuan Wang and Zhaoxin Yu and Eric Hanchen Jiang and Song-Chun Zhu and Zixia Jia and Ying Nian Wu and Zilong Zheng},
  eprint        = {2505.13308},
  journal       = {arXiv preprint arXiv:2505.13308},
  primaryclass  = {cs.LG},
  title         = {Seek in the Dark: Reasoning via Test-Time Instance-Level Policy Gradient in Latent Space},
  url           = {https://arxiv.org/abs/2505.13308},
  year          = {2026}
}

@article{liu2026thoughtbubblesunsupervisedmethodparallel,
  archiveprefix = {arXiv},
  author        = {Houjun Liu and Shikhar Murty and Christopher D. Manning and Róbert Csordás},
  eprint        = {2510.00219},
  journal       = {arXiv preprint arXiv:2510.00219},
  primaryclass  = {cs.LG},
  title         = {Thoughtbubbles: an Unsupervised Method for Parallel Thinking in Latent Space},
  url           = {https://arxiv.org/abs/2510.00219},
  year          = {2026}
}

@article{chen2026thinkingdeeperlongerdepthrecurrent,
  archiveprefix = {arXiv},
  author        = {Hung-Hsuan Chen},
  eprint        = {2603.21676},
  journal       = {arXiv preprint arXiv:2603.21676},
  primaryclass  = {cs.LG},
  title         = {Thinking Deeper, Not Longer: Depth-Recurrent Transformers for Compositional Generalization},
  url           = {https://arxiv.org/abs/2603.21676},
  year          = {2026}
}

@article{moosa2026understandingdynamiccomputeallocation,
  archiveprefix = {arXiv},
  author        = {Ibraheem Muhammad Moosa and Suhas Lohit and Ye Wang and Moitreya Chatterjee and Wenpeng Yin},
  eprint        = {2602.08864},
  journal       = {arXiv preprint arXiv:2602.08864},
  primaryclass  = {cs.CL},
  title         = {Understanding Dynamic Compute Allocation in Recurrent Transformers},
  url           = {https://arxiv.org/abs/2602.08864},
  year          = {2026}
}

@article{amos2026latentreasoningsupervisedthinking,
  archiveprefix = {arXiv},
  author        = {Ido Amos and Avi Caciularu and Mor Geva and Amir Globerson and Jonathan Herzig and Lior Shani and Idan Szpektor},
  eprint        = {2602.08332},
  journal       = {arXiv preprint arXiv:2602.08332},
  primaryclass  = {cs.CL},
  title         = {Latent Reasoning with Supervised Thinking States},
  url           = {https://arxiv.org/abs/2602.08332},
  year          = {2026}
}

@inproceedings{pfau2024lets,
  author    = {Jacob Pfau and William Merrill and Samuel R. Bowman},
  booktitle = {COLM},
  title     = {Let{\textquoteright}s Think Dot by Dot: Hidden computation in transformer language models},
  url       = {https://openreview.net/forum?id=NikbrdtYvG},
  year      = {2024}
}

@article{cheng2024compressedchainthoughtefficient,
  archiveprefix = {arXiv},
  author        = {Jeffrey Cheng and Benjamin Van Durme},
  eprint        = {2412.13171},
  journal       = {arXiv preprint arXiv:2412.13171},
  primaryclass  = {cs.CL},
  title         = {Compressed Chain of Thought: Efficient Reasoning Through Dense Representations},
  url           = {https://arxiv.org/abs/2412.13171},
  year          = {2024}
}

@article{zou2026capabilitiesfundamentallimitslatent,
  archiveprefix = {arXiv},
  author        = {Jiaxuan Zou and Yaozhong Xiong and Yong Liu},
  eprint        = {2602.01148},
  journal       = {arXiv preprint arXiv:2602.01148},
  primaryclass  = {cs.AI},
  title         = {Capabilities and Fundamental Limits of Latent Chain-of-Thought},
  url           = {https://arxiv.org/abs/2602.01148},
  year          = {2026}
}

@article{wang2026latentchainofthoughtplanningdecoupling,
  archiveprefix = {arXiv},
  author        = {Jiecong Wang and Hao Peng and Chunyang Liu},
  eprint        = {2601.21358},
  journal       = {arXiv preprint arXiv:2601.21358},
  primaryclass  = {cs.AI},
  title         = {Latent Chain-of-Thought as Planning: Decoupling Reasoning from Verbalization},
  url           = {https://arxiv.org/abs/2601.21358},
  year          = {2026}
}

@article{li2025implicitreasoninglargelanguage,
  archiveprefix = {arXiv},
  author        = {Jindong Li and Yali Fu and Li Fan and Jiahong Liu and Yao Shu and Chengwei Qin and Menglin Yang and Irwin King and Rex Ying},
  eprint        = {2509.02350},
  journal       = {arXiv preprint arXiv:2509.02350},
  primaryclass  = {cs.CL},
  title         = {Implicit Reasoning in Large Language Models: A Comprehensive Survey},
  url           = {https://arxiv.org/abs/2509.02350},
  year          = {2025}
}

@article{deng2026llmlatentreasoningchain,
  archiveprefix = {arXiv},
  author        = {Jingcheng Deng and Liang Pang and Zihao Wei and Shicheng Xu and Zenghao Duan and Kun Xu and Yang Song and Huawei Shen and Xueqi Cheng},
  eprint        = {2510.15522},
  journal       = {arXiv preprint arXiv:2510.15522},
  primaryclass  = {cs.CL},
  title         = {LLM Latent Reasoning as Chain of Superposition},
  url           = {https://arxiv.org/abs/2510.15522},
  year          = {2026}
}

@article{geiping2025scalingup,
  archiveprefix = {arXiv},
  author        = {Jonas Geiping and Sean McLeish and Neel Jain and John Kirchenbauer and Siddharth Singh and Brian R. Bartoldson and Bhavya Kailkhura and Abhinav Bhatele and Tom Goldstein},
  eprint        = {2502.05171},
  journal       = {arXiv preprint arXiv:2502.05171},
  primaryclass  = {cs.LG},
  title         = {Scaling up Test-Time Compute with Latent Reasoning: A Recurrent Depth Approach},
  url           = {https://arxiv.org/abs/2502.05171},
  year          = {2025}
}

@inproceedings{geiping2025efficient,
  author    = {Jonas Geiping and Xinyu Yang and Guinan Su},
  booktitle = {NeurIPS 2025 Workshop on Efficient Reasoning},
  title     = {Efficient Parallel Samplers for Recurrent-Depth Models and Their Connections to Diffusion Language Models},
  url       = {https://openreview.net/forum?id=nA5IRfAfbn},
  year      = {2025}
}

@article{knupp2026depthrecurrentattentionmixturesgiving,
  archiveprefix = {arXiv},
  author        = {Jonas Knupp and Jan Hendrik Metzen and Jeremias Bohn and Georg Groh and Kristian Kersting},
  eprint        = {2601.21582},
  journal       = {arXiv preprint arXiv:2601.21582},
  primaryclass  = {cs.AI},
  title         = {Depth-Recurrent Attention Mixtures: Giving Latent Reasoning the Attention it Deserves},
  url           = {https://arxiv.org/abs/2601.21582},
  year          = {2026}
}

@article{li2026chainthoughtcompressiontheoritical,
  archiveprefix = {arXiv},
  author        = {Juncai Li and Ru Li and Yuxiang Zhou and Boxiang Ma and Jeff Z. Pan},
  eprint        = {2601.21576},
  journal       = {arXiv preprint arXiv:2601.21576},
  primaryclass  = {cs.AI},
  title         = {Chain Of Thought Compression: A Theoritical Analysis},
  url           = {https://arxiv.org/abs/2601.21576},
  year          = {2026}
}

@article{cobbe2021trainingverifierssolvemath,
  archiveprefix = {arXiv},
  author        = {Karl Cobbe and Vineet Kosaraju and Mohammad Bavarian and Mark Chen and Heewoo Jun and Lukasz Kaiser and Matthias Plappert and Jerry Tworek and Jacob Hilton and Reiichiro Nakano and Christopher Hesse and John Schulman},
  eprint        = {2110.14168},
  journal       = {arXiv preprint arXiv:2110.14168},
  primaryclass  = {cs.LG},
  title         = {Training Verifiers to Solve Math Word Problems},
  url           = {https://arxiv.org/abs/2110.14168},
  year          = {2021}
}

@article{ng2024loopneuralnetworksparameter,
  archiveprefix = {arXiv},
  author        = {Kei-Sing Ng and Qingchen Wang},
  eprint        = {2409.14199},
  journal       = {arXiv preprint arXiv:2409.14199},
  primaryclass  = {cs.AI},
  title         = {Loop Neural Networks for Parameter Sharing},
  url           = {https://arxiv.org/abs/2409.14199},
  year          = {2024}
}

@article{Xu2025AFC,
  author  = {Kevin Xu and Issei Sato},
  journal = {ArXiv},
  title   = {A Formal Comparison Between Chain-of-Thought and Latent Thought},
  url     = {https://api.semanticscholar.org/CorpusID:281681404},
  volume  = {abs/2509.25239},
  year    = {2025}
}

@article{xu2025cotloopformalcomparison,
  archiveprefix = {arXiv},
  author        = {Kevin Xu and Issei Sato},
  eprint        = {2505.19245},
  journal       = {arXiv preprint arXiv:2505.19245},
  primaryclass  = {cs.LG},
  title         = {To {CoT} or To Loop? {A} Formal Comparison Between Chain-of-Thought and Looped Transformers},
  url           = {https://arxiv.org/abs/2505.19245},
  year          = {2025}
}

@article{lian2025threadweaveradaptivethreadingefficient,
  archiveprefix = {arXiv},
  author        = {Long Lian and Sida Wang and Felix Juefei-Xu and Tsu-Jui Fu and Xiuyu Li and Adam Yala and Trevor Darrell and Alane Suhr and Yuandong Tian and Xi Victoria Lin},
  eprint        = {2512.07843},
  journal       = {arXiv preprint arXiv:2512.07843},
  primaryclass  = {cs.LG},
  title         = {ThreadWeaver: Adaptive Threading for Efficient Parallel Reasoning in Language Models},
  url           = {https://arxiv.org/abs/2512.07843},
  year          = {2025}
}

@inproceedings{liu2025deliberation,
  author    = {Luyang Liu and Jonas Pfeiffer and Jiaxing Wu and Jun Xie and Arthur Szlam},
  booktitle = {Forty-second International Conference on Machine Learning},
  title     = {Deliberation in Latent Space via Differentiable Cache Augmentation},
  url       = {https://openreview.net/forum?id=IaUJl5RCOu},
  year      = {2025}
}

@article{liu2025tidar,
  title={Tidar: Think in diffusion, talk in autoregression},
  author={Liu, Jingyu and Dong, Xin and Ye, Zhifan and Mehta, Rishabh and Fu, Yonggan and Singh, Vartika and Kautz, Jan and Zhang, Ce and Molchanov, Pavlo},
  journal={arXiv preprint arXiv:2511.08923},
  year={2025}
}

@inproceedings{maile2026dynamicparameterreuse,
  author    = {Maile, Kaitlin and Sacramento, João},
  booktitle = {ICLR Blogposts 2026},
  date      = {April 27, 2026},
  note      = {https://iclr-blogposts.github.io/2026/blog/2026/recur-refine-reason/},
  title     = {Dynamic Parameter Reuse Augments Reasoning via Latent Chain of Thought},
  url       = {https://iclr-blogposts.github.io/2026/blog/2026/recur-refine-reason/},
  year      = {2026}
}

@inproceedings{rizvi-martel2026the,
  author    = {Michael Rizvi-Martel and Marius Mosbach},
  booktitle = {Workshop on Latent {\&} Implicit Thinking {\textendash} Going Beyond CoT Reasoning},
  title     = {The Illusion of Superposition in Latent CoT via Soft Thinking},
  url       = {https://openreview.net/forum?id=FvPx9Nzvnw},
  year      = {2026}
}

@article{dehghani2019universaltransformers,
  archiveprefix = {arXiv},
  author        = {Mostafa Dehghani and Stephan Gouws and Oriol Vinyals and Jakob Uszkoreit and Łukasz Kaiser},
  eprint        = {1807.03819},
  journal       = {arXiv preprint arXiv:1807.03819},
  primaryclass  = {cs.CL},
  title         = {Universal Transformers},
  url           = {https://arxiv.org/abs/1807.03819},
  year          = {2019}
}

@inproceedings{saunshi2025reasoninglatentthoughtspower,
  author    = {Nikunj Saunshi and Nishanth Dikkala and Zhiyuan Li and Sanjiv Kumar and Sashank J. Reddi},
  booktitle = {The Thirteenth International Conference on Learning Representations},
  title     = {Reasoning with Latent Thoughts: On the Power of Looped Transformers},
  url       = {https://openreview.net/forum?id=din0lGfZFd},
  year      = {2025}
}

@inproceedings{nowak-etal-2024-representational,
  address   = {Bangkok, Thailand},
  author    = {Nowak, Franz  and
               Svete, Anej  and
               Butoi, Alexandra  and
               Cotterell, Ryan},
  booktitle = {Proceedings of the 62nd Annual Meeting of the Association for Computational Linguistics (Volume 1: Long Papers)},
  doi       = {10.18653/v1/2024.acl-long.676},
  editor    = {Ku, Lun-Wei  and
               Martins, Andre  and
               Srikumar, Vivek},
  month     = aug,
  pages     = {12510--12548},
  publisher = {Association for Computational Linguistics},
  title     = {On the Representational Capacity of Neural Language Models with Chain-of-Thought Reasoning},
  url       = {https://aclanthology.org/2024.acl-long.676/},
  year      = {2024}
}

@article{openai2026openaio1card,
  archiveprefix = {arXiv},
  author        = {OpenAI},
  eprint        = {2412.16720},
  journal       = {arXiv preprint arXiv:2412.16720},
  primaryclass  = {cs.AI},
  title         = {{OpenAI o1} System Card},
  url           = {https://arxiv.org/abs/2412.16720},
  year          = {2026}
}

@inproceedings{csordas2025do,
  author    = {R{\'o}bert Csord{\'a}s and Christopher D Manning and Christopher Potts},
  booktitle = {The Thirty-ninth Annual Conference on Neural Information Processing Systems},
  title     = {Do Language Models Use Their Depth Efficiently?},
  url       = {https://openreview.net/forum?id=Kz6eUL86XP},
  year      = {2025}
}

@inproceedings{csordas2024moeut,
  author    = {R{\'o}bert Csord{\'a}s and Kazuki Irie and J{\"u}rgen Schmidhuber and Christopher Potts and Christopher D Manning},
  booktitle = {The Thirty-eighth Annual Conference on Neural Information Processing Systems},
  title     = {Mo{EUT}: Mixture-of-Experts Universal Transformers},
  url       = {https://openreview.net/forum?id=ZxVrkm7Bjl},
  year      = {2024}
}

@article{fu2026selarselectivelatentreasoning,
  archiveprefix = {arXiv},
  author        = {Renyu Fu and Guibo Luo},
  eprint        = {2604.08299},
  journal       = {arXiv preprint arXiv:2604.08299},
  primaryclass  = {cs.CL},
  title         = {SeLaR: Selective Latent Reasoning in Large Language Models},
  url           = {https://arxiv.org/abs/2604.08299},
  year          = {2026}
}

@article{zhu2025surveylatentreasoning,
  archiveprefix = {arXiv},
  author        = {Rui-Jie Zhu and Tianhao Peng and Tianhao Cheng and Xingwei Qu and Jinfa Huang and Dawei Zhu and Hao Wang and Kaiwen Xue and Xuanliang Zhang and Yong Shan and Tianle Cai and Taylor Kergan and Assel Kembay and Andrew Smith and Chenghua Lin and Binh Nguyen and Yuqi Pan and Yuhong Chou and Zefan Cai and Zhenhe Wu and Yongchi Zhao and Tianyu Liu and Jian Yang and Wangchunshu Zhou and Chujie Zheng and Chongxuan Li and Yuyin Zhou and Zhoujun Li and Zhaoxiang Zhang and Jiaheng Liu and Ge Zhang and Wenhao Huang and Jason Eshraghian},
  eprint        = {2507.06203},
  journal       = {arXiv preprint arXiv:2507.06203},
  primaryclass  = {cs.CL},
  title         = {A Survey on Latent Reasoning},
  url           = {https://arxiv.org/abs/2507.06203},
  year          = {2025}
}

@article{zhu2025scalinglatentreasoninglooped,
  archiveprefix = {arXiv},
  author        = {Rui-Jie Zhu and Zixuan Wang and Kai Hua and Tianyu Zhang and Ziniu Li and Haoran Que and Boyi Wei and Zixin Wen and Fan Yin and He Xing and Lu Li and Jiajun Shi and Kaijing Ma and Shanda Li and Taylor Kergan and Andrew Smith and Xingwei Qu and Mude Hui and Bohong Wu and Qiyang Min and Hongzhi Huang and Xun Zhou and Wei Ye and Jiaheng Liu and Jian Yang and Yunfeng Shi and Chenghua Lin and Enduo Zhao and Tianle Cai and Ge Zhang and Wenhao Huang and Yoshua Bengio and Jason Eshraghian},
  eprint        = {2510.25741},
  journal       = {arXiv preprint arXiv:2510.25741},
  primaryclass  = {cs.CL},
  title         = {Scaling Latent Reasoning via Looped Language Models},
  url           = {https://arxiv.org/abs/2510.25741},
  year          = {2025}
}

@article{you2026paralleltesttimescalinglatent,
  archiveprefix = {arXiv},
  author        = {Runyang You and Yongqi Li and Meng Liu and Wenjie Wang and Liqiang Nie and Wenjie Li},
  eprint        = {2510.07745},
  journal       = {arXiv preprint arXiv:2510.07745},
  primaryclass  = {cs.CL},
  title         = {Parallel Test-Time Scaling for Latent Reasoning Models},
  url           = {https://arxiv.org/abs/2510.07745},
  year          = {2026}
}

@inproceedings{goyal2024think,
  author    = {Sachin Goyal and Ziwei Ji and Ankit Singh Rawat and Aditya Krishna Menon and Sanjiv Kumar and Vaishnavh Nagarajan},
  booktitle = {The Twelfth International Conference on Learning Representations},
  title     = {Think before you speak: Training Language Models With Pause Tokens},
  url       = {https://openreview.net/forum?id=ph04CRkPdC},
  year      = {2024}
}

@inproceedings{bae2025relaxed,
  author    = {Sangmin Bae and Adam Fisch and Hrayr Harutyunyan and Ziwei Ji and Seungyeon Kim and Tal Schuster},
  booktitle = {The Thirteenth International Conference on Learning Representations},
  title     = {Relaxed Recursive Transformers: Effective Parameter Sharing with Layer-wise Lo{RA}},
  url       = {https://openreview.net/forum?id=WwpYSOkkCt},
  year      = {2025}
}

@inproceedings{bae2025mixtureofrecursions,
  author    = {Sangmin Bae and Yujin Kim and Reza Bayat and Sungnyun Kim and Jiyoun Ha and Tal Schuster and Adam Fisch and Hrayr Harutyunyan and Ziwei Ji and Aaron Courville and Se-Young Yun},
  booktitle = {The Thirty-ninth Annual Conference on Neural Information Processing Systems},
  title     = {Mixture-of-Recursions: Learning Dynamic Recursive Depths for Adaptive Token-Level Computation},
  url       = {https://openreview.net/forum?id=QuqsEIVWIG},
  year      = {2025}
}

@article{mcleish2025teachingpretrainedlanguagemodels,
  archiveprefix = {arXiv},
  author        = {Sean McLeish and Ang Li and John Kirchenbauer and Dayal Singh Kalra and Brian R. Bartoldson and Bhavya Kailkhura and Avi Schwarzschild and Jonas Geiping and Tom Goldstein and Micah Goldblum},
  eprint        = {2511.07384},
  journal       = {arXiv preprint arXiv:2511.07384},
  primaryclass  = {cs.CL},
  title         = {Teaching Pretrained Language Models to Think Deeper with Retrofitted Recurrence},
  url           = {https://arxiv.org/abs/2511.07384},
  year          = {2025}
}

@article{jerad2026contextfreerecognitiontransformers,
  archiveprefix = {arXiv},
  author        = {Selim Jerad and Anej Svete and Sophie Hao and Ryan Cotterell and William Merrill},
  eprint        = {2601.01754},
  journal       = {arXiv preprint arXiv:2601.01754},
  primaryclass  = {cs.CL},
  title         = {Context-Free Recognition with Transformers},
  url           = {https://arxiv.org/abs/2601.01754},
  year          = {2026}
}

@inproceedings{shen-etal-2025-codi,
  address   = {Suzhou, China},
  author    = {Shen, Zhenyi  and
               Yan, Hanqi  and
               Zhang, Linhai  and
               Hu, Zhanghao  and
               Du, Yali  and
               He, Yulan},
  booktitle = {Proceedings of the 2025 Conference on Empirical Methods in Natural Language Processing},
  doi       = {10.18653/v1/2025.emnlp-main.36},
  editor    = {Christodoulopoulos, Christos  and
               Chakraborty, Tanmoy  and
               Rose, Carolyn  and
               Peng, Violet},
  isbn      = {979-8-89176-332-6},
  month     = nov,
  pages     = {677--693},
  publisher = {Association for Computational Linguistics},
  title     = {{CODI}: Compressing Chain-of-Thought into Continuous Space via Self-Distillation},
  url       = {https://aclanthology.org/2025.emnlp-main.36/},
  year      = {2025}
}

@inproceedings{hao2025training,
  author    = {Shibo Hao and Sainbayar Sukhbaatar and DiJia Su and Xian Li and Zhiting Hu and Jason E Weston and Yuandong Tian},
  booktitle = {Second Conference on Language Modeling},
  title     = {Training Large Language Models to Reason in a Continuous Latent Space},
  url       = {https://openreview.net/forum?id=Itxz7S4Ip3},
  year      = {2025}
}

@article{song2026adaponderlmgatedponderinglanguage,
  archiveprefix = {arXiv},
  author        = {Shixiang Song and He Li and Zitong Wang and Boyi Zeng and Feichen Song and Yixuan Wang and Zhiqin John Xu and Ziwei He and Zhouhan Lin},
  eprint        = {2603.01914},
  journal       = {arXiv preprint arXiv:2603.01914},
  primaryclass  = {cs.CL},
  title         = {AdaPonderLM: Gated Pondering Language Models with Token-Wise Adaptive Depth},
  url           = {https://arxiv.org/abs/2603.01914},
  year          = {2026}
}

@article{feng2025efficientreasoningmodelssurvey,
  archiveprefix = {arXiv},
  author        = {Sicheng Feng and Gongfan Fang and Xinyin Ma and Xinchao Wang},
  eprint        = {2504.10903},
  journal       = {arXiv preprint arXiv:2504.10903},
  primaryclass  = {cs.CL},
  title         = {Efficient Reasoning Models: A Survey},
  url           = {https://arxiv.org/abs/2504.10903},
  year          = {2025}
}

@article{fu2025thinkathardselectivelatentiterations,
  archiveprefix = {arXiv},
  author        = {Tianyu Fu and Yichen You and Zekai Chen and Guohao Dai and Huazhong Yang and Yu Wang},
  eprint        = {2511.08577},
  journal       = {arXiv preprint arXiv:2511.08577},
  primaryclass  = {cs.CL},
  title         = {Think-at-Hard: Selective Latent Iterations to Improve Reasoning Language Models},
  url           = {https://arxiv.org/abs/2511.08577},
  year          = {2025}
}

@inproceedings{10.5555/3600270.3602070,
  address   = {Red Hook, NY, USA},
  articleno = {1800},
  author    = {Wei, Jason and Wang, Xuezhi and Schuurmans, Dale and Bosma, Maarten and Ichter, Brian and Xia, Fei and Chi, Ed H. and Le, Quoc V. and Zhou, Denny},
  booktitle = {Proceedings of the 36th International Conference on Neural Information Processing Systems},
  isbn      = {9781713871088},
  location  = {New Orleans, LA, USA},
  numpages  = {14},
  publisher = {Curran Associates Inc.},
  series    = {NIPS '22},
  title     = {Chain-of-thought prompting elicits reasoning in large language models},
  year      = {2022}
}

@inproceedings{ye2026thinking,
  author    = {Wengao Ye and Yan Liang and Lianlei Shan},
  booktitle = {The Fourteenth International Conference on Learning Representations},
  title     = {Thinking on the Fly: Test-Time Reasoning Enhancement via Latent Thought Policy Optimization},
  url       = {https://openreview.net/forum?id=r1WEQzkCQv},
  year      = {2026}
}

@inproceedings{tan2025think,
  author    = {Wenhui Tan and Jiaze Li and Jianzhong Ju and Zhenbo Luo and Ruihua Song and Jian Luan},
  booktitle = {The Thirty-ninth Annual Conference on Neural Information Processing Systems},
  title     = {Think Silently, Think Fast: Dynamic Latent Compression of {LLM} Reasoning Chains},
  url       = {https://openreview.net/forum?id=AQsko3PPUe},
  year      = {2025}
}

@inproceedings{merrill2025littledepthgoeslong,
  author    = {William Merrill and Ashish Sabharwal},
  booktitle = {The Thirty-ninth Annual Conference on Neural Information Processing Systems},
  title     = {A Little Depth Goes a Long Way: The Expressive Power of Log-Depth Transformers},
  url       = {https://openreview.net/forum?id=5pHfYe10iX},
  year      = {2025}
}

@inproceedings{merrill2025exactexpressivepowertransformers,
  author    = {William Merrill and Ashish Sabharwal},
  booktitle = {The Thirty-ninth Annual Conference on Neural Information Processing Systems},
  title     = {Exact Expressive Power of Transformers with Padding},
  url       = {https://openreview.net/forum?id=O1abxStFcy},
  year      = {2025}
}

@inproceedings{merrill2024the,
  author    = {William Merrill and Ashish Sabharwal},
  booktitle = {The Twelfth International Conference on Learning Representations},
  title     = {The Expressive Power of Transformers with Chain of Thought},
  url       = {https://openreview.net/forum?id=NjNGlPh8Wh},
  year      = {2024}
}

@inproceedings{wu-etal-2025-parallel,
  address   = {Suzhou, China},
  author    = {Wu, Haoyi  and
               Teng, Zhihao  and
               Tu, Kewei},
  booktitle = {Proceedings of the 2025 Conference on Empirical Methods in Natural Language Processing},
  doi       = {10.18653/v1/2025.emnlp-main.47},
  editor    = {Christodoulopoulos, Christos  and
               Chakraborty, Tanmoy  and
               Rose, Carolyn  and
               Peng, Violet},
  isbn      = {979-8-89176-332-6},
  month     = nov,
  pages     = {914--926},
  publisher = {Association for Computational Linguistics},
  title     = {Parallel Continuous Chain-of-Thought with {J}acobi Iteration},
  url       = {https://aclanthology.org/2025.emnlp-main.47/},
  year      = {2025}
}

@article{wei2025simcotsupervisedimplicitchainofthought,
  archiveprefix = {arXiv},
  author        = {Xilin Wei and Xiaoran Liu and Yuhang Zang and Xiaoyi Dong and Yuhang Cao and Jiaqi Wang and Xipeng Qiu and Dahua Lin},
  eprint        = {2509.20317},
  journal       = {arXiv preprint arXiv:2509.20317},
  primaryclass  = {cs.CL},
  title         = {{SIM-CoT}: {S}upervised Implicit Chain-of-Thought},
  url           = {https://arxiv.org/abs/2509.20317},
  year          = {2025}
}

@article{chen2025reasoninglanguagecomprehensivesurvey,
  archiveprefix = {arXiv},
  author        = {Xinghao Chen and Anhao Zhao and Heming Xia and Xuan Lu and Hanlin Wang and Yanjun Chen and Wei Zhang and Jian Wang and Wenjie Li and Xiaoyu Shen},
  eprint        = {2505.16782},
  journal       = {arXiv preprint arXiv:2505.16782},
  primaryclass  = {cs.CL},
  title         = {Reasoning Beyond Language: A Comprehensive Survey on Latent Chain-of-Thought Reasoning},
  url           = {https://arxiv.org/abs/2505.16782},
  year          = {2025}
}

@article{yu2026latentspacefoundationevolution,
  archiveprefix = {arXiv},
  author        = {Xinlei Yu and Zhangquan Chen and Yongbo He and Tianyu Fu and Cheng Yang and Chengming Xu and Yue Ma and Xiaobin Hu and Zhe Cao and Jie Xu and Guibin Zhang and Jiale Tao and Jiayi Zhang and Siyuan Ma and Kaituo Feng and Haojie Huang and Youxing Li and Ronghao Chen and Huacan Wang and Chenglin Wu and Zikun Su and Xiaogang Xu and Kelu Yao and Kun Wang and Chen Gao and Yue Liao and Ruqi Huang and Tao Jin and Cheng Tan and Jiangning Zhang and Wenqi Ren and Yanwei Fu and Yong Liu and Yu Wang and Xiangyu Yue and Yu-Gang Jiang and Shuicheng Yan},
  eprint        = {2604.02029},
  journal       = {arXiv preprint arXiv:2604.02029},
  primaryclass  = {cs.AI},
  title         = {The Latent Space: Foundation, Evolution, Mechanism, Ability, and Outlook},
  url           = {https://arxiv.org/abs/2604.02029},
  year          = {2026}
}

@inproceedings{wang2024guiding,
  author    = {Xinyi Wang and Lucas Caccia and Oleksiy Ostapenko and Xingdi Yuan and William Yang Wang and Alessandro Sordoni},
  booktitle = {First Conference on Language Modeling},
  title     = {Guiding Language Model Reasoning with Planning Tokens},
  url       = {https://openreview.net/forum?id=wi9IffRhVM},
  year      = {2024}
}

@article{koishekenov2025encodethinkdecodescaling,
  archiveprefix = {arXiv},
  author        = {Yeskendir Koishekenov and Aldo Lipani and Nicola Cancedda},
  eprint        = {2510.07358},
  journal       = {arXiv preprint arXiv:2510.07358},
  primaryclass  = {cs.LG},
  title         = {Encode, Think, Decode: Scaling test-time reasoning with recursive latent thoughts},
  url           = {https://arxiv.org/abs/2510.07358},
  year          = {2025}
}

@article{xu2025softcottesttimescalingsoft,
  archiveprefix = {arXiv},
  author        = {Yige Xu and Xu Guo and Zhiwei Zeng and Chunyan Miao},
  eprint        = {2505.11484},
  journal       = {arXiv preprint arXiv:2505.11484},
  primaryclass  = {cs.CL},
  title         = {SoftCoT++: Test-Time Scaling with Soft Chain-of-Thought Reasoning},
  url           = {https://arxiv.org/abs/2505.11484},
  year          = {2025}
}

@inproceedings{fan2025looped,
  author    = {Ying Fan and Yilun Du and Kannan Ramchandran and Kangwook Lee},
  booktitle = {The Thirteenth International Conference on Learning Representations},
  title     = {Looped Transformers for Length Generalization},
  url       = {https://openreview.net/forum?id=2edigk8yoU},
  year      = {2025}
}

@article{cui2026latentreasoningmethodsperform,
  archiveprefix = {arXiv},
  author        = {Yingqian Cui and Zhenwei Dai and Bing He and Zhan Shi and Hui Liu and Rui Sun and Zhiji Liu and Yue Xing and Jiliang Tang and Benoit Dumoulin},
  eprint        = {2602.22441},
  journal       = {arXiv preprint arXiv:2602.22441},
  primaryclass  = {cs.AI},
  title         = {How Do Latent Reasoning Methods Perform Under Weak and Strong Supervision?},
  url           = {https://arxiv.org/abs/2602.22441},
  year          = {2026}
}

@inproceedings{he2025semcot,
  author    = {Yinhan He and Wendy Zheng and Yaochen Zhu and Zaiyi Zheng and Lin Su and Sriram Vasudevan and Qi Guo and Liangjie Hong and Jundong Li},
  booktitle = {The Thirty-ninth Annual Conference on Neural Information Processing Systems},
  title     = {SemCoT: Accelerating Chain-of-Thought Reasoning through Semantically-Aligned Implicit Tokens},
  url       = {https://openreview.net/forum?id=1ZuzFUMtx6},
  year      = {2025}
}

@article{chu2026spotspanlevelpauseofthoughtefficient,
  archiveprefix = {arXiv},
  author        = {Yunlong Chu and Minglai Shao and Yuhang Liu and Bing Hao and Yumeng Lin and Jialu Wang and Ruijie Wang},
  eprint        = {2603.06222},
  journal       = {arXiv preprint arXiv:2603.06222},
  primaryclass  = {cs.CL},
  title         = {SPOT: Span-level Pause-of-Thought for Efficient and Interpretable Latent Reasoning in Large Language Models},
  url           = {https://arxiv.org/abs/2603.06222},
  year          = {2026}
}

@article{zhang2025latenttokensthinkcausal,
  archiveprefix = {arXiv},
  author        = {Yuyi Zhang and Boyu Tang and Tianjie Ju and Sufeng Duan and Gongshen Liu},
  eprint        = {2512.21711},
  journal       = {arXiv preprint arXiv:2512.21711},
  primaryclass  = {cs.CL},
  title         = {Do Latent Tokens Think? A Causal and Adversarial Analysis of Chain-of-Continuous-Thought},
  url           = {https://arxiv.org/abs/2512.21711},
  year          = {2025}
}

@article{chen2025thinkconsistentlyreasonefficiently,
  archiveprefix = {arXiv},
  author        = {Zhikang Chen and Sen Cui and Deheng Ye and Yu Zhang and Yatao Bian and Tingting Zhu},
  eprint        = {2511.07124},
  journal       = {arXiv preprint arXiv:2511.07124},
  primaryclass  = {cs.CL},
  title         = {Think Consistently, Reason Efficiently: Energy-Based Calibration for Implicit Chain-of-Thought},
  url           = {https://arxiv.org/abs/2511.07124},
  year          = {2025}
}

@inproceedings{li2024chain,
  author    = {Zhiyuan Li and Hong Liu and Denny Zhou and Tengyu Ma},
  booktitle = {The Twelfth International Conference on Learning Representations},
  title     = {Chain of Thought Empowers Transformers to Solve Inherently Serial Problems},
  url       = {https://openreview.net/forum?id=3EWTEy9MTM},
  year      = {2024}
}

@article{radford2019language,
  title={Language models are unsupervised multitask learners},
  author={Radford, Alec and Wu, Jeffrey and Child, Rewon and Luan, David and Amodei, Dario and Sutskever, Ilya and others}
}

@article{grattafiori2024llama,
  title={The llama 3 herd of models},
  author={Grattafiori, Aaron and Dubey, Abhimanyu and Jauhri, Abhinav and Pandey, Abhinav and Kadian, Abhishek and Al-Dahle, Ahmad and Letman, Aiesha and Mathur, Akhil and Schelten, Alan and Vaughan, Alex and others},
  journal={arXiv preprint arXiv:2407.21783},
  year={2024}
}
